\documentclass[preprint,12pt]{elsarticle}

\usepackage{caption}
\usepackage{amssymb}
\usepackage{amsthm}
\usepackage{lineno}
\usepackage{bm}
\usepackage[colorlinks=true, pdfauthor=author]{hyperref}
\usepackage{wrapfig}
\usepackage{booktabs} 
\usepackage{algorithm}
\usepackage{algorithmic}
\usepackage{graphicx}
\usepackage{amsmath}
\usepackage{amssymb}
\usepackage{amsfonts}
\usepackage{pifont}
\usepackage{multirow}
\usepackage{graphicx}
\usepackage{mathrsfs}
\usepackage{colortbl}
\usepackage{subfigure}
\usepackage{indentfirst}
\usepackage{tcolorbox}
\usepackage{ulem}
\usepackage{url}
\usepackage{placeins}
\usepackage{dblfloatfix}
\usepackage{verbatim}
\usepackage{rotating}
\usepackage{adjustbox}
\usepackage{wrapfig}
\usepackage{setspace}
\usepackage{array}
\usepackage{booktabs}
\newcolumntype{L}[1]{>{\raggedright\arraybackslash}p{#1}}

\graphicspath{{./images/}}
\usepackage[table]{xcolor}

\journal{arXiv}
\begin{document}

\begin{frontmatter}

\title{Tabular Imbalanced Learning: A Survey, Benchmark, and Practical Guide}

\author[first]{Ruizhe Liu}
\ead{2307401023@stu.suda.edu.cn}

\author[first]{Jiaqi Luo\corref{cor}}
\ead{jqluo@suda.edu.cn}
\affiliation[first]{organization={School of Mathematical Sciences, Soochow University},
            addressline={No.1 Shizi Street}, 
            city={Suzhou},
            postcode={215006}, 
            state={Jiangsu Province},
            country={China}}
\cortext[cor]{Corresponding author}

\begin{abstract}

Imbalanced learning remains a fundamental challenge in tabular data applications. Despite decades of research and numerous proposed methods, there is still limited systematic understanding of how different imbalance-handling strategies perform across diverse data regimes and computational constraints, making practical method selection difficult.
In this work, we provide a systematic survey of tabular imbalanced learning and introduce \textbf{T}abular \textbf{I}mbalanced \textbf{L}earning \textbf{Bench}mark (\textbf{TILBench}), a large-scale empirical benchmark for evaluating existing methods. We first organize imbalanced learning approaches into a unified taxonomy and then benchmark more than 40 representative methods across 57 tabular datasets under a standardized evaluation protocol, examining overall predictive performance, sensitivity to dataset characteristics, and computational scalability.
Our results show that no single method consistently dominates across all settings. Instead, the effectiveness of different strategies depends strongly on dataset regimes and computational constraints, highlighting the need for regime-aware method selection. Based on these findings, we provide practical recommendations for selecting imbalanced learning methods under different data conditions and identify open challenges and promising directions for future research.

\end{abstract}




\end{frontmatter}

\section{Introduction}
\label{s.intro}
\begin{figure*}[!ht]
    \centering
    \includegraphics[width=0.95\textwidth]{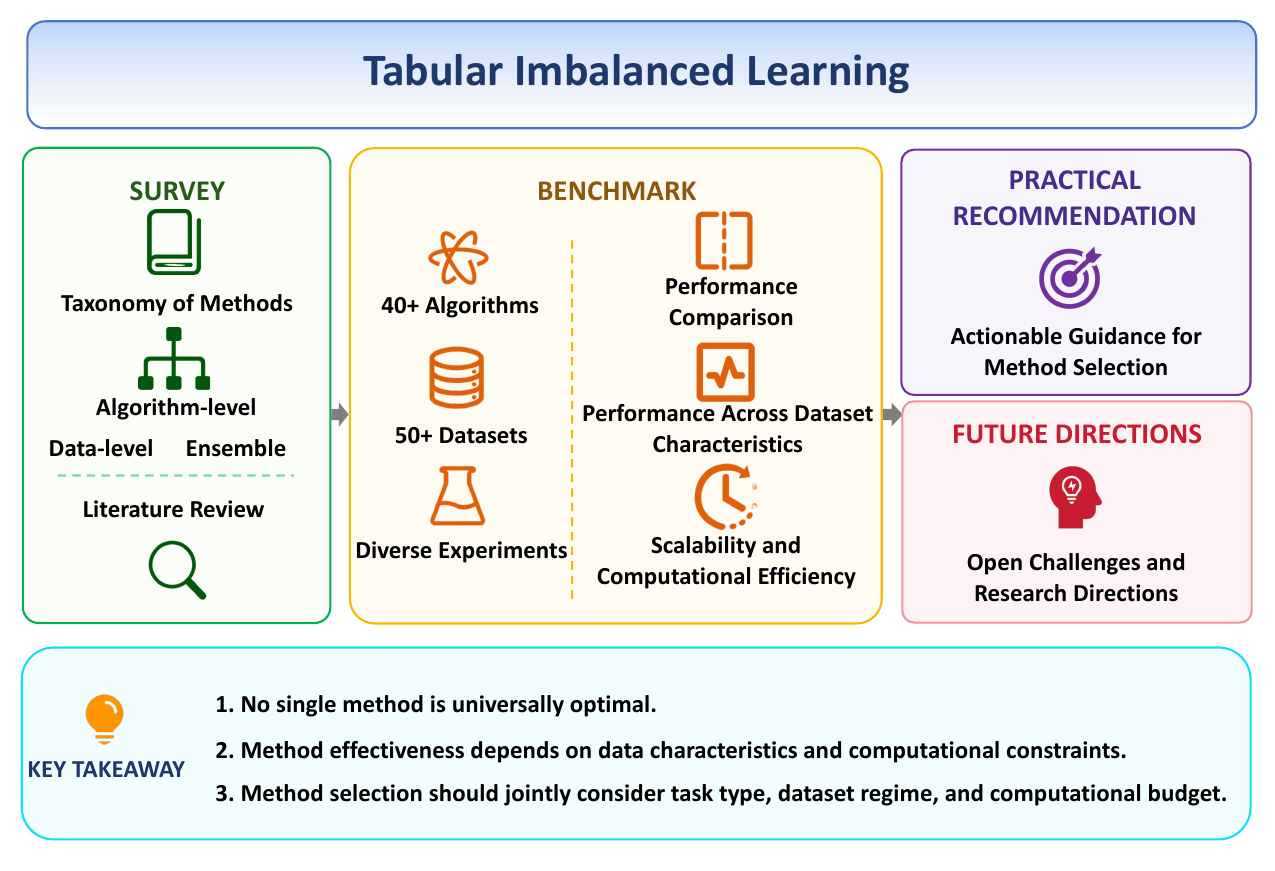}
    \caption{
    Overview of our study. We provide a systematic taxonomy of tabular imbalanced learning methods and then introduce TILBench, which evaluates more than 40 representative methods across 57 tabular datasets. The empirical analyses examine overall performance, behavior across dataset characteristics, and computational scalability, leading to practical recommendations and future research directions.
    }
\end{figure*}

Imbalanced learning is a fundamental challenge in tabular machine learning~\cite{he2009learning,haixiang2017learning,fernandez2018learning,krawczyk2016learning}, where minority cases are rare but often critical. Such settings arise in many real-world applications, including fraud detection~\cite{zhu2020optimizing}, medical diagnosis~\cite{santos2022decision}, and fault diagnosis~\cite{zhang2018imbalanced}. In these scenarios, conventional learning algorithms tend to favor the majority class, resulting in poor recognition of minority instances despite their greater practical importance.

To address this issue, a wide range of imbalanced learning methods have been proposed, including data-level techniques~\cite{smith2014instance,chawla2002smote,han2005borderline}, algorithm-level approaches~\cite{luo2025improving,xu2020imbalanced,liu2022focal}, and ensemble-based strategies~\cite{liu2020self,karakoulas1998optimizing,viola2001fast}. These methods address class imbalance through fundamentally different mechanisms, such as modifying the data distribution, adapting the learning process, or combining multiple learners. Consequently, their strengths and limitations are expected to vary across different data characteristics and computational settings, making systematic empirical comparison essential.

Existing studies mainly fall into two categories: surveys and empirical benchmarks. Surveys provide comprehensive taxonomies and conceptual understanding of imbalanced learning methods~\cite{he2009learning,haixiang2017learning,fernandez2018learning}, but generally lack systematic empirical validation across diverse tabular settings. Empirical benchmarks, on the other hand, compare algorithm performance but often focus on a limited subset of method families, datasets, or application domains~\cite{khan2024review,kovacs2019empirical,liu2025climb}. More importantly, existing empirical studies~\cite{liu2025climb,aguiar2024survey} provide limited evidence on how method effectiveness changes with dataset characteristics and computational scale. Consequently, there remains a gap between understanding \textit{how} different imbalance-handling strategies operate and knowing \textit{when} they are effective in practice.

In this work, we provide a systematic survey of tabular imbalanced learning and introduce \textbf{T}abular \textbf{I}mbalanced \textbf{L}earning \textbf{Bench}mark (\textbf{TILBench}), a large-scale empirical benchmark for evaluating existing methods. We first organize existing approaches into a unified taxonomy covering data-level, algorithm-level, and ensemble-based methods, summarizing their mechanisms, representative algorithms, strengths, and limitations. TILBench then evaluates more than 40 representative methods across 57 diverse tabular datasets. By connecting methodological characteristics with empirical behavior, our study further translates the benchmark findings into practical recommendations and identifies open challenges for future research.

The empirical benchmark addresses three complementary research questions. First, we examine how individual algorithms and method families compare in overall predictive performance. Second, we investigate how their effectiveness varies with sample size, feature dimensionality, imbalance severity, and missing values. Third, we evaluate computational scalability under increasing sample size, feature dimensionality, and class number. Rather than seeking a single globally optimal method, these analyses aim to identify when different imbalance-handling strategies are most effective and what trade-offs arise under different tabular data conditions.

The main contributions of this work are summarized as follows:
\begin{itemize}
\item We provide a systematic survey and unified taxonomy of tabular imbalanced learning methods, summarizing the mechanisms, representative algorithms, advantages, and limitations of the three major methodological families.

\item We introduce TILBench, a large-scale empirical benchmark covering more than 40 representative methods and 57 tabular datasets, enabling systematic comparison across method families.

\item We investigate how method effectiveness varies across dataset characteristics, including sample size, feature dimensionality, imbalance severity, and missing values, revealing regime-dependent performance patterns beyond a single global ranking.

\item We analyze computational scalability under increasing sample size, feature dimensionality, and class number, characterizing the practical cost of different imbalance-handling strategies as data complexity increase.

\item By connecting methodological mechanisms with empirical evidence, we provide regime-aware recommendations for practical method selection and identify open challenges and promising directions for future research on tabular imbalanced learning.
\end{itemize}

\section{Taxonomy and Literature Review of Tabular Imbalanced Learning}
\label{s:prelim}

\subsection{Problem Setting and Method Taxonomy}
\label{s:work}

Imbalanced learning considers classification problems in which the class distribution is highly skewed and standard learning algorithms tend to favor majority classes. In tabular applications, this challenge is further affected by heterogeneous feature spaces, limited minority samples, missing values, and differences in dataset scale and dimensionality.

\begin{figure}[!ht]
    \centering
    \includegraphics[width=0.9\linewidth]{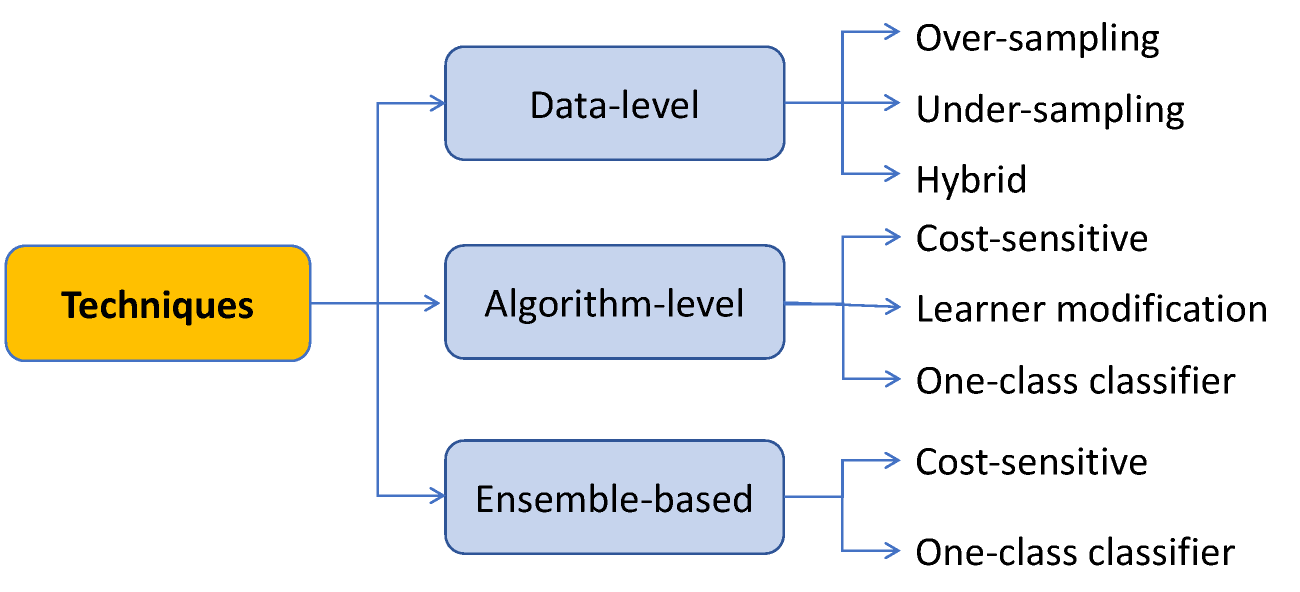}
    \caption{Taxonomy of tabular imbalanced learning methods. Existing approaches are organized into data-level, algorithm-level, and ensemble-based families according to where imbalance is addressed in the learning pipeline.}
    \label{f.class}
\end{figure}

Existing approaches can be broadly categorized into three families according to where imbalance is addressed in the learning pipeline: \textbf{data-level methods}, which modify the empirical training distribution; \textbf{algorithm-level methods}, which incorporate imbalance awareness into the learning process; and \textbf{ensemble-based methods}, which combine multiple imbalance-aware learners. An overview of this taxonomy is illustrated in Fig.~\ref{f.class}.

\subsubsection{Data-Level Methods}

Data-level methods, also referred to as \textit{resampling methods}, address class imbalance by modifying the training data distribution prior to model learning while leaving the underlying classifier unchanged. Since they operate as preprocessing procedures, they are largely model-agnostic and can therefore be integrated with a wide range of learning algorithms. Existing methods can be broadly categorized into oversampling, undersampling, and hybrid approaches.

\paragraph{Oversampling}

Oversampling methods increase the representation of minority classes by duplicating existing samples or generating synthetic minority instances. Random oversampling is the simplest strategy, but repeatedly duplicating minority samples may increase overfitting and contribute little new information. To alleviate this issue, synthetic oversampling methods generate new samples in the feature space. The seminal Synthetic Minority Over-sampling Technique (SMOTE)~\cite{chawla2002smote} synthesizes minority instances by interpolating between neighboring minority samples and has become the foundation of many subsequent methods. Representative extensions include Borderline-SMOTE~\cite{han2005borderline}, which concentrates sample generation near decision boundaries, ADASYN~\cite{he2008adasyn}, which adaptively allocates more synthetic samples to difficult regions, and KMeansSMOTE~\cite{douzas2018improving}, which incorporates clustering information before interpolation. Numerous additional variants have also been proposed to improve sample diversity and robustness under different data characteristics~\cite{kovacs2019empirical,gazzah2008new,saez2015smote,lee2015over}.

More recently, generative models~\cite{goodfellow2020generative,ho2020denoising} have emerged as an alternative to conventional oversampling. Instead of synthesizing samples through local interpolation, these methods learn the underlying data distribution and generate more realistic minority instances. Representative approaches include GAN-based models such as CTGAN~\cite{xu2019modeling}, CTAB-GAN~\cite{zhao2021ctab}, CTAB-GAN+~\cite{zhao2024ctab}, and Conditional Wasserstein GAN~\cite{engelmann2021conditional}, as well as diffusion-based generators including TabDDPM~\cite{kotelnikov2023tabddpm}, TabDiff~\cite{shi2025tabdiff}, and Mixed-Type Diffusion Models~\cite{zhang2024mixed}. Although generative models often produce higher-quality synthetic samples, they typically require substantially greater computational resources and larger training datasets.

\paragraph{Undersampling}

Undersampling methods reduce class imbalance by removing majority-class samples. Compared with oversampling, these methods avoid generating artificial data and often reduce training cost, but aggressive sample removal may discard informative observations and distort the decision boundary. Random undersampling provides the simplest solution, whereas more sophisticated methods attempt to preserve representative majority samples. TomekLinks~\cite{tomek1976two} removes majority samples located near class boundaries, while NearMiss~\cite{mani2003knn}, Cluster Centroids~\cite{lin2017clustering}, Condensed Nearest Neighbor, and Edited Nearest Neighbor~\cite{hart1968condensed,laurikkala2001improving,wilson2007asymptotic} retain informative samples according to distance-based or neighborhood-based criteria. Recent studies further improve sample selection by incorporating data characteristics into the undersampling process. For example, UBMD~\cite{sun2024undersampling} selects majority samples according to minority density, while Learning-To-Rank Undersampling (LTRUS) formulates sample selection as a learning-to-rank problem~\cite{feng2024improving}.

\paragraph{Hybrid Methods}

Hybrid methods combine oversampling and undersampling to simultaneously enrich minority classes and remove noisy or overlapping majority samples. Classical approaches include SMOTETomek~\cite{batista2003balancing}, which combines SMOTE with TomekLinks, and SMOTEENN~\cite{batista2004study}, which integrates SMOTE with Edited Nearest Neighbors. More recently, hybrid strategies have also incorporated deep generative models. For example, CTGAN-ENN~\cite{adiputra2024ctgan} combines CTGAN-based oversampling with Edited Nearest Neighbors, while HD-GNNS~\cite{liu2024hybrid} integrates GAN-based sample generation with natural-neighbor search to simultaneously alleviate class imbalance and class overlap. Although hybrid methods often achieve improved data quality, they generally introduce additional computational overhead and hyperparameter tuning complexity.

\subsubsection{Algorithm-Level Methods}

Algorithm-level methods address class imbalance by modifying the learning process rather than altering the training data. Unlike resampling methods, these approaches preserve the original data distribution and incorporate imbalance awareness directly into the learning procedure. Existing methods can be broadly categorized into cost-sensitive learning, learner modification, and one-class classification.

\paragraph{Cost-sensitive Learning}

Cost-sensitive learning addresses class imbalance by assigning different penalties to different classes or types of prediction error. By increasing the cost associated with minority-class misclassification, these methods encourage the learner to pay more attention to underrepresented classes while preserving the original training distribution. 
Representative approaches include class-weighted classifiers and wrapper methods such as MetaCost~\cite{domingos1999metacost}, which estimates class probabilities and relabels training instances according to expected misclassification costs. Cost-sensitive variants have also been developed for many widely used learning algorithms, including decision trees~\cite{ling2004decision,ting2002instance}, support vector machines~\cite{datta2015near}, random forests~\cite{breiman2001random}, and gradient-boosted decision trees such as XGBoost~\cite{chen2016xgboost} and LightGBM~\cite{ke2017lightgbm}.

\paragraph{Learner Modification}

Learner modification methods explicitly alter the internal learning mechanism of a model to reduce majority-class bias. Unlike cost-sensitive learning, which modifies the optimization objective, these methods redesign components of the learning algorithm itself. Representative examples include kernel modification for support vector machines~\cite{wu2005kba}, imbalance-aware split criteria for decision trees~\cite{hoens2012building}, and other model-specific strategies that improve minority-class discrimination~\cite{cano2013weighted,diamantini2008bayes,yu2018active}. Because learner modification exploits the characteristics of individual models, it can provide more targeted improvements than generic weighting strategies. However, these methods are typically model-dependent and often require substantial algorithmic redesign, limiting their general applicability.

Recent years have witnessed substantial progress in imbalance-aware loss functions, particularly in deep learning. Representative examples include Weighted Cross-Entropy~\cite{sun2009classification}, Focal Loss~\cite{lin2017focal}, Class-Balanced Loss~\cite{cui2019class}, and Asymmetric Loss~\cite{ridnik2021asymmetric}, together with numerous variants designed to improve optimization under severe imbalance~\cite{cao2019learning,fernando2021dynamically,wu2020distribution,wang2020imbalance,tian2023synergetic}. While these loss functions have achieved considerable success in neural networks, many cannot be directly incorporated into gradient-boosted decision trees because GBDT optimization relies on tractable first- and second-order derivatives. Recent studies~\cite{luo2025improving,luo2025robust} address this challenge by extending modern GBDT frameworks to support a broader class of loss functions.

\paragraph{One-Class Classification}

One-class classification treats minority-class recognition as an anomaly detection problem by learning the distribution of a target class and identifying observations outside this distribution as anomalies~\cite{manevitz2001one}. This formulation is particularly suitable for extremely imbalanced applications, where abnormal events are rare and highly diverse. Representative methods include one-class support vector machines~\cite{manevitz2001one}, as well as approaches that combine one-class learning with ensemble methods, oversampling, or instance selection~\cite{czarnowski2022weighted}. Closely related anomaly detection frameworks have also been widely adopted for identifying minority and majority instances~\cite{han2022adbench}. Although one-class methods are effective when only normal data are readily available, their performance often deteriorates under substantial class overlap, noisy observations, or inappropriate kernel and threshold selection.

\subsubsection{Ensemble-Based Methods}

Ensemble-based methods integrate imbalance-aware learning strategies into ensemble learning frameworks. By combining multiple base learners, these methods exploit both sample diversity and model diversity to improve robustness under skewed class distributions. Depending on how imbalance is incorporated, existing approaches can be broadly categorized into data-level ensembles and algorithm-level ensembles.

\paragraph{Data-Level Ensembles}

Data-level ensemble methods construct multiple balanced training subsets through resampling and train an individual base learner on each subset. The final prediction is obtained by aggregating the outputs of all learners, thereby reducing variance while improving minority-class recognition.
Representative approaches include bagging-based methods, such as UnderBagging~\cite{maclin1997empirical} and SMOTEBagging~\cite{wang2009diversity}, which generate multiple resampled datasets and combine the resulting classifiers. Boosting-based methods further integrate resampling into iterative ensemble learning. For example, SMOTEBoost~\cite{chawla2003smoteboost} combines SMOTE with AdaBoost, while RUSBoost~\cite{seiffert2009rusboost} incorporates random undersampling into the boosting procedure. Another important family consists of undersampling ensembles, including EasyEnsemble~\cite{liu2008exploratory}, BalanceCascade~\cite{liu2008exploratory}, and the recently proposed Self-Paced Ensemble~\cite{liu2022predicting}. 
By combining resampling with ensemble learning, these methods generally achieve stronger minority-class performance and improved robustness than single resampling techniques. However, repeated resampling and training multiple base learners inevitably increase computational cost, particularly for large-scale or high-dimensional datasets.

\paragraph{Algorithm-Level Ensembles}

Algorithm-level ensemble methods incorporate imbalance awareness directly into the ensemble construction process without modifying the original data distribution. These approaches redesign the learning objectives or cost-sensitive updates to emphasize minority classes throughout ensemble optimization.
Representative methods include AdaCost~\cite{fan1999adacost}, which extends AdaBoost by introducing class-dependent misclassification costs into both sample-weight updating and ensemble combination. Related approaches such as AdaUBoost~\cite{karakoulas1998optimizing} and AsymBoost~\cite{viola2001fast} similarly modify the boosting procedure to assign greater importance to minority samples during training. Compared with data-level ensembles, these methods preserve the original data distribution and avoid repeated resampling, but they often require careful calibration of class costs or update rules and are generally more model-specific.

\subsubsection{Unified Perspective and Key Differences}

Although the three methodological families differ substantially in their implementation, they can be viewed from a unified perspective according to the stage at which class imbalance is addressed. Data-level methods modify the empirical training distribution before learning, algorithm-level methods incorporate imbalance awareness directly into the learning objective or optimization procedure, and ensemble-based methods improve robustness by combining multiple imbalance-aware learners.

These different intervention mechanisms naturally lead to different trade-offs. Data-level methods are generally flexible and model-agnostic, but may distort the original data distribution or introduce noisy synthetic samples. Algorithm-level methods preserve the training data and offer direct control over learning bias, but are often model-dependent and require algorithm-specific design. Ensemble-based methods typically achieve stronger robustness and predictive performance by exploiting model diversity, at the expense of increased computational cost due to repeated training and model aggregation.

These key characteristics are summarized in Table~\ref{T.key_differences}. They also motivate the empirical benchmark presented in the following sections, which investigates how the effectiveness of each methodological family varies across different data characteristics and computational settings.

\begin{table}[!t]
\centering
\caption{Key differences among imbalanced learning method families.}
\label{T.key_differences}
\begin{adjustbox}{width=\linewidth}
\begin{tabular}{llll}
\toprule[2pt]
\textbf{Aspect}
& \textbf{Data-level}
& \textbf{Algorithm-level}
& \textbf{Ensemble-based} \\
\midrule[1.5pt]
Mechanism
& Data distribution
& Learning process
& Model aggregation \\
\hline
Operation
& Resampling
& Learner modification
& Repeated training \\
\hline
Advantage
& Flexibility
& Bias control
& Robustness \\
\hline
Limitation
& Distribution distortion
& Model dependence
& Computational cost \\
\bottomrule[2pt]
\end{tabular}
\end{adjustbox}
\end{table}

\subsection{Existing Surveys and Benchmarks}

The literature on imbalanced learning can be broadly divided into surveys and empirical benchmarks. Existing surveys provide valuable summaries of methodological developments, while benchmarks focus on comparing algorithm performance under specific experimental settings.

Foundational surveys and monographs~\cite{he2009learning,haixiang2017learning,fernandez2018learning} provide comprehensive overviews of major methodological paradigms, including resampling, cost-sensitive learning, learner modification, and ensemble methods. More recent reviews concentrate on particular research directions, such as data-level techniques~\cite{nikpour2026comprehensive}, cost-sensitive learning~\cite{khan2018cost,araf2024cost}, and support vector machine-based approaches~\cite{rezvani2023broad}. These surveys provide useful conceptual organization and summarize the evolution of existing methods, but generally do not offer large-scale empirical comparisons across diverse tabular learning scenarios.

A complementary line of research focuses on empirical benchmarking. Representative studies evaluate oversampling methods~\cite{kovacs2019empirical}, data augmentation and ensemble strategies~\cite{khan2024review}, or imbalance-aware loss functions~\cite{luo2025improving}. More comprehensive benchmarks~\cite{liu2025climb,aguiar2024survey} compare multiple categories of methods, but still cover only a subset of the available methodological families, with learner modification approaches being relatively underrepresented. In addition, several benchmarks have been developed for specific application domains, including business~\cite{zhu2018benchmarking}, finance~\cite{xiao2021impact}, and education~\cite{wongvorachan2023comparison}. Although these studies provide valuable domain-specific evidence, their conclusions may not generalize to broader tabular learning settings.

Overall, existing benchmarks remain limited in two important aspects. First, their coverage of methods and datasets is often incomplete, preventing a unified comparison across the three major methodological families. Second, most studies primarily report average predictive performance and provide limited analysis of how method effectiveness varies with dataset characteristics, such as sample size, feature dimensionality, imbalance severity, and missing-value patterns, or with computational scalability. Consequently, there remains a need for a unified benchmark that connects the taxonomy of existing methods with systematic empirical evidence and provides practical guidance for regime-aware method selection.

\begin{table}[t]
\centering
\caption{Comparison between TILBench and representative surveys and empirical benchmarks on tabular imbalanced learning.}
\label{T.comparison}
\renewcommand{\arraystretch}{1.15}
\begin{adjustbox}{width=\linewidth}
\begin{tabular}{l|cc|ccc|cc}
\toprule[2pt]
\multirow{2}{*}{\textbf{Work}}
& \multicolumn{2}{c|}{\textbf{Scope}}
& \multicolumn{3}{c|}{\textbf{Method Coverage}}
& \multicolumn{2}{c}{\textbf{Analysis}} \\
\cline{2-8}
&
Survey &
Benchmark &
Data-level &
Algorithm-level &
Ensemble &
Data-Regime &
Practical Guide \\
\midrule

\cite{khan2024review}
& \checkmark & \ding{55}
& \checkmark & \ding{55} & \checkmark
& \ding{55} & \ding{55} \\

\cite{kovacs2019empirical}
& \ding{55} & \checkmark
& \checkmark & \ding{55} & \ding{55}
& \ding{55} & \ding{55} \\

\cite{liu2025climb}
& \ding{55} & \checkmark
& \checkmark & \ding{55} & \checkmark
& \ding{55} & \ding{55} \\

\cite{aguiar2024survey}
& \checkmark & \checkmark
& \checkmark & \checkmark & \ding{55}
& \ding{55} & \ding{55} \\

\cite{nikpour2026comprehensive}
& \checkmark & \ding{55}
& \checkmark & \ding{55} & \ding{55}
& \ding{55} & \ding{55} \\

\cite{araf2024cost}
& \checkmark & \ding{55}
& \ding{55} & \checkmark & \ding{55}
& \ding{55} & \ding{55} \\

\midrule

\textbf{Ours}
& \checkmark & \checkmark
& \checkmark & \checkmark & \checkmark
& \checkmark & \checkmark \\

\bottomrule[2pt]
\end{tabular}
\end{adjustbox}
\end{table}


\section{Benchmark Design}
\label{s:design}

\subsection{Scope and Objectives}

We conduct a large-scale empirical study to evaluate tabular imbalanced learning methods under diverse data conditions. Rather than seeking a single universally best algorithm, the benchmark is designed to identify how the relative effectiveness of different method families changes across data regimes and computational settings.

The study addresses the following research questions:

\begin{itemize}
    \item \textbf{RQ1:} How do different families of imbalanced learning methods compare in overall predictive performance?

    \item \textbf{RQ2:} How does the effectiveness of imbalanced learning methods vary with sample size, feature dimensionality, imbalance severity, and missing values?

    \item \textbf{RQ3:} How do different imbalanced learning methods compare in computational scalability and efficiency as data complexity increase?
\end{itemize}

To answer these questions, we construct a controlled evaluation pipeline with consistent data splits, hyperparameter tuning budgets, evaluation metrics, and implementation settings across methods.

\subsection{Datasets}

Our evaluation includes 57 tabular datasets, consisting of 34 binary classification tasks and 23 multi-class classification tasks. Among them, 7 datasets contain natural missing values. The datasets are collected from OpenML~\cite{vanschoren2014openml} and imbalanced-learn\footnote{\url{https://imbalanced-learn.org/stable/index.html}}.

The dataset collection is designed to cover a broad range of tabular imbalanced learning conditions rather than a single application domain. The datasets vary substantially in sample size, feature dimensionality, imbalance severity, number of classes, and missing-value patterns. This diversity allows us to examine both overall predictive performance and the dependence of method effectiveness on specific dataset characteristics. Detailed dataset statistics and regime assignments are reported in Table~\ref{T.data} in \ref{a.imp}.

\subsection{Evaluated Methods}

We evaluate more than 40 representative methods spanning the three families introduced in Section~\ref{s:prelim}: data-level, algorithm-level, and ensemble-based approaches. Methods are selected according to three criteria: methodological representativeness, use in prior imbalanced learning research, and availability of reproducible implementations. The benchmark includes both classical approaches and more recent variants, allowing comparisons across method families as well as within-family designs.

We use XGBoost~\cite{chen2016xgboost} as the reference learner because Gradient Boosting Decision Trees consistently demonstrate strong performance on tabular data~\cite{borisov2022deep,grinsztajn2022tree,luo2024ncart}. XGBoost is also used as the base classifier for all data-level methods. This design controls for differences in the underlying learner and allows performance changes to be attributed primarily to the resampling strategy.

\paragraph{Data-level methods}
The evaluated undersampling methods include TomekLinks~\cite{tomek1976two}, EditedNearestNeighbors~\cite{wilson2007asymptotic}, NeighborhoodCleaningRule~\cite{laurikkala2001improving}, InstanceHardnessThreshold~\cite{smith2014instance}, ClusterCentroids~\cite{lin2017clustering}, CondensedNearestNeighbor~\cite{hart1968condensed}, AllKNN~\cite{tomek1976experiment}, NearMiss~\cite{mani2003knn}, and One-SidedSelection~\cite{kubat1997addressing}.
The evaluated oversampling methods include SMOTE~\cite{chawla2002smote}, BorderlineSMOTE~\cite{han2005borderline}, PolyfitSMOTE~\cite{gazzah2008new}, SMOTEIPF~\cite{saez2015smote}, Lee~\cite{lee2015over}, and SMOBD~\cite{cao2011applying}. The hybrid methods are SMOTEENN~\cite{batista2004study} and SMOTETomek~\cite{batista2003balancing}.

\paragraph{Algorithm-level methods}
The cost-sensitive methods include XGBoostCost, LogisticRegressionCost, DecisionTreeCost, and RandomForestCost. These models incorporate class-dependent weights while preserving the original data distribution.
The learner modification methods combine XGBoost with imbalance-aware loss functions, including Asymmetric Loss (XGBoostASL)~\cite{ridnik2021asymmetric}, Asymmetric Cross-Entropy (XGBoostACE)~\cite{luo2025improving}, Asymmetric Weighted Cross-Entropy (XGBoostAWE)~\cite{luo2025improving}, Focal Loss (XGBoostFL)~\cite{lin2017focal}, Weighted Cross-Entropy (XGBoostWCE)~\cite{sun2009classification}, and Class-Balanced Cross-Entropy (XGBoostCBE)~\cite{cui2019class}.

\paragraph{Ensemble-based methods}
The data-level ensemble methods include SelfPacedEnsemble~\cite{liu2020self}, BalanceCascadeEnsemble~\cite{liu2008exploratory}, BalancedRandomForest~\cite{chen2004using}, EasyEnsemble~\cite{liu2008exploratory}, RUSBoost~\cite{seiffert2009rusboost}, UnderBagging~\cite{maclin1997empirical}, SMOTEBoost~\cite{chawla2003smoteboost}, OverBoost~\cite{chawla2003smoteboost}, OverBagging~\cite{maclin1997empirical}, and SMOTEBagging~\cite{wang2009diversity}.
The algorithm-level ensemble methods include AdaUBoost~\cite{karakoulas1998optimizing}, AdaCost~\cite{fan1999adacost}, and AsymBoost~\cite{viola2001fast}.

\subsection{Evaluation Protocol}

To ensure fair and reproducible evaluation, we adopt a standardized protocol across all datasets and methods. Each dataset is divided into 80\% training and 20\% test data using stratified sampling. Within the training split, 10\% of the samples are reserved for validation and hyperparameter selection.

Hyperparameters are optimized using Optuna~\cite{akiba2019optuna} with 20 trials per method. 
After hyperparameter selection, each model is retrained using the full training split and evaluated on the held-out test set. 
This entire process is repeated with five random seeds, and we report the mean and standard deviation of the macro-averaged \textbf{F1-score} and \textbf{G-mean}. For $C$ classes, F1-score is defined as
\begin{equation}
\mathrm{F1\text{-}score}=\frac{1}{C}\sum_{c=1}^{C}\frac{2P_cR_c}{P_c+R_c},
\end{equation}
where $P_c$ and $R_c$ denote the precision and recall of class $c$, respectively. G-mean is computed with macro averaging as
\begin{equation}
\mathrm{G\text{-}mean}=\sqrt{\bar{R}\,\bar{S}},
\end{equation}
where $\bar{R}$ and $\bar{S}$ denote the macro-averaged sensitivity and specificity across classes, respectively.

All preprocessing and resampling operations are fitted exclusively on the training data to prevent information leakage. The same random splits are shared across methods within each seed, ensuring paired comparisons. Data-level methods use XGBoost as the common downstream classifier, while algorithm-level and ensemble methods retain their designated learning mechanisms.

All experiments are conducted on a workstation equipped with an Intel Core i9-14900HX CPU and 48 GB RAM. Hyperparameter search spaces, software packages, implementation sources, and method-specific settings are reported in Table~\ref{T.hyperparam} and Table~\ref{T.python} in \ref{a.imp}.

\section{Results and Analyses}
\label{s:results}

\subsection{Overall Performance Comparison}

We evaluate the overall effectiveness of imbalanced learning methods from two complementary perspectives: the global performance of individual algorithms and the aggregate behavior of method families.

\subsubsection{Global Ranking}


We first compare individual methods across all datasets. Table~\ref{top_ten_f1} and~\ref{top_ten_gm} report the ten best-performing methods in terms of F1-score and G-mean for binary and multi-class classification, respectively. Complete results are provided in Tables~\ref{T.all_binary} and~\ref{T.all_multi} in \ref{a.sup}.

Several observations emerge from the results. 
\textit{First, method rankings are sensitive to the evaluation metric.}
For binary classification, SelfPacedEnsemble achieves the highest F1-score, whereas UnderBagging obtains the highest G-mean despite ranking considerably lower in F1-score. Similar discrepancies are observed for other methods, highlighting the need for complementary metrics when evaluating imbalanced learning methods.

\textit{Second, different strategies are favored in binary and multi-class classification.}
Binary tasks favor ensemble-based and algorithm-level approaches, with SelfPacedEnsemble leading in F1-score and several XGBoost-based loss modifications performing strongly. In multi-class tasks, data-level methods become more competitive. SMOTE achieves the highest F1-score, and several resampling variants rank highly under both metrics. This suggests that the effectiveness of a strategy depends on the classification setting rather than being consistent across task types.

\textit{Third, a strong base learner remains highly competitive, particularly in multi-class classification.}
The original XGBoost ranks highly under both metrics on multi-class tasks, with only small gaps from several imbalance-aware variants. In binary classification, however, it is substantially less competitive, and explicit imbalance handling provides clearer improvements. 
This suggests that when the base learner is already strong, additional imbalance handling may provide only limited gains.

\begin{table}[!ht]
\centering
\caption{Top 10 performing methods ranked by F1-score for binary and multi-class tasks.}
\label{top_ten_f1}
\begin{adjustbox}{width=\textwidth}
\begin{tabular}{c|c|c||c|c|c}
\toprule
\textbf{Rank} & \textbf{Method} & \textbf{F1-score} & \textbf{Rank} & \textbf{Method} & \textbf{F1-score} \\
\midrule
\multicolumn{3}{c}{Binary} & \multicolumn{3}{c}{Multi-class}\\
\midrule
1&SelfPacedEnsemble & 73.88 $\pm$ 4.01 &1 &SMOTE & 75.84 $\pm$ 1.45\\
2&XGBoostWCE &72.96 $\pm$ 4.91 &2 &XGBoostCost & 75.83 $\pm$ 1.57\\
3&XGBoostASL &72.62 $\pm$ 5.48 &3 &SMOTETomek &75.49 $\pm$ 1.45\\
4&XGBoostAWE &72.54 $\pm$ 4.57 &4 &BorderlineSMOTE & 75.49 $\pm$ 1.71\\
5&BalanceCascadeEnsemble &72.51 $\pm$ 4.55 &5 &Lee &75.47 $\pm$ 1.52\\
6&XGBoostACE &72.07 $\pm$ 6.16 &6 &Tomek & 75.02 $\pm$ 1.67\\
7&SMOTEBagging &70.93 $\pm$ 4.73 &7 &XGBoostFL &74.69 $\pm$ 1.76\\
8&XGBoostLee &70.53 $\pm$ 5.97 &8 &XGBoostCBE &74.69 $\pm$ 1.64\\
9&XGBoostFL &70.14 $\pm$ 5.84 &9 &XGBoost &74.63 $\pm$ 1.61\\
10&XGBoostCost &69.90 $\pm$ 4.90 &10 &XGBoostWCE &74.61 $\pm$ 1.31\\
\bottomrule
\end{tabular}
\end{adjustbox}
\end{table}

\begin{table}[!ht]
\centering
\caption{Top 10 performing methods ranked by G-mean for binary and multi-class tasks.}
\label{top_ten_gm}
\begin{adjustbox}{width=\textwidth}
\begin{tabular}{c|c|c||c|c|c}
\toprule
\textbf{Rank} & \textbf{Method} & \textbf{G-mean} & \textbf{Rank} & \textbf{Method} & \textbf{G-mean} \\
\midrule
\multicolumn{3}{c}{Binary} & \multicolumn{3}{c}{Multi-class}\\
\midrule
1 & UnderBagging & 89.32 $\pm$ 2.12 & 1 & XGBoostCost & 83.91 $\pm$ 1.16\\
2 & BalancedRandomForest & 88.34 $\pm$ 2.38 & 2 &SMOTE & 83.63 $\pm$ 1.00\\
3 & SMOTEENN & 87.27 $\pm$ 3.21 & 3 & SMOTETomek & 83.44 $\pm$ 0.94\\
4 & BalanceCascadeEnsemble & 87.20 $\pm$ 2.72 & 4 & BorderlineSMOTE & 83.23 $\pm$ 1.26\\
5 & AdaUCost & 86.74 $\pm$ 3.43 & 5 & Lee & 83.08 $\pm$ 1.04\\
6 & EasyEnsemble & 86.70 $\pm$ 2.40 & 6 & TomekLinks & 82.60 $\pm$ 1.29\\
7 & XGBoostCost & 86.22 $\pm$ 3.39 & 7 & SMOTEIPF & 82.49 $\pm$ 1.11\\
8 & SelfPacedEnsemble & 85.97 $\pm$ 3.14 & 8 & XGBoostFL & 82.15 $\pm$ 1.29\\
9 & SMOTEIPF & 85.00 $\pm$ 3.58 & 9 & XGBoostCBE & 82.15 $\pm$ 1.19\\
10 & SMOTETomek & 84.92 $\pm$ 3.99 & 10 & XGBoostWCE & 82.12 $\pm$ 0.95\\
\bottomrule
\end{tabular}
\end{adjustbox}
\end{table}

\subsubsection{Family-level Aggregation}

\begin{table}[!ht]
\centering
\caption{Family-level performance comparison across binary and multi-class tasks in terms of F1-score and G-mean. We report average performance, standard deviation, and extreme values (max/min) within each family.}
\label{family}
\begin{adjustbox}{width=\linewidth}
\begin{tabular}{c|c|c|c|c|c}
\toprule[2pt]
\multicolumn{2}{c|}{} & \textbf{\large XGBoost} & \textbf{\large Data-level} & \textbf{\large Alg.-level} & \textbf{\large Ensemble}\\
\midrule[1.5pt]
\multicolumn{6}{c}{F1-score}\\
\midrule[1.5pt]
\multirow{5}{*}{Binary} & Rank & --& 3& 1&2 \\
\cline{2-6}
& Avg. &60.92 &60.78 & 68.43&63.95 \\
\cline{2-6}
& Std. &-- &9.84 &5.75 & 6.89\\
\cline{2-6}
& Max &60.92 &70.53 &72.96 &73.88 \\
\cline{2-6}
& Min &60.92 &36.71  &55.30 & 50.79\\
\midrule
\multirow{5}{*}{Multi-class} & Rank & --& 2&1 &3 \\
\cline{2-6}
& Avg. &74.63 &70.51& 71.23& 63.61\\
\cline{2-6}
& Std. &-- &5.67 & 5.00&7.36 \\
\cline{2-6}
& Max &74.63 &75.84 &75.83 &73.11 \\
\cline{2-6}
& Min &74.63 &58.47 & 60.11 & 50.44 \\
\midrule[1.5pt]
\multicolumn{6}{c}{G-mean}\\
\midrule[1.5pt]
\multirow{5}{*}{Binary} & Rank &-- & 3&1 & 2\\
\cline{2-6}
& Avg. & 67.17&77.04 & 82.52& 82.15\\
\cline{2-6}
& Std. & --& 6.94& 3.15 &6.29 \\
\cline{2-6}
& Max & 67.17&87.27 &86.22& 89.32\\
\cline{2-6}
& Min & 67.17&65.95 &77.05 &71.41 \\
\midrule
\multirow{5}{*}{Multi-class} & Rank & --&1 & 2&3 \\
\cline{2-6}
& Avg. &82.09 &80.57 &80.46& 76.08\\
\cline{2-6}
& Std. &-- & 2.69&2.78 &5.08\\
\cline{2-6}
& Max &82.09 &83.63 & 83.91 &81.92 \\
\cline{2-6}
& Min &82.09 &75.78 & 74.12 & 66.77\\
\bottomrule[2pt]
\end{tabular}
\end{adjustbox}
\end{table}

To complement the individual rankings, we further aggregate methods into data-level, algorithm-level, and ensemble-based families. Table~\ref{family} summarizes their average performance and within-family variation under both F1-score and G-mean.

The family-level results reveal three main conclusions. \textit{First, algorithm-level methods provide the strongest and most consistent family-level performance in binary classification.}
They achieve the highest average F1-score and G-mean with the smallest variation across methods. Ensemble-based methods are also competitive, particularly in G-mean, while data-level methods perform worse on average. These results establish algorithm-level approaches as the most reliable family for binary tasks.

\textit{Second, the gap between method families becomes smaller in multi-class classification.} Algorithm-level methods achieve the best average F1-score, while data-level methods obtain a slightly higher average G-mean. The small gap between these two families suggests that multi-class imbalance can benefit from both learning-process modification and data distribution adjustment. Meanwhile, ensemble-based methods show weaker average performance in this setting, indicating that their advantages in binary classification do not necessarily transfer to multi-class tasks.

\textit{Third, strong individual methods do not necessarily imply a strong method family.}
Ensemble-based methods contain several top-performing algorithms, particularly in binary classification, yet their family-level averages and larger variation reveal considerable inconsistency among individual approaches. Similar variation is observed within data-level methods. Therefore, family-level performance reflects the general reliability of a strategy, while individual algorithms can deviate substantially from that trend.

\begin{tcolorbox}[colback=gray!10, colframe=black!50, title=\textbf{Summary for RQ1: Overall Performance Comparison}]
Algorithm-level methods provide the most stable overall performance across datasets. Ensemble-based methods are highly competitive in binary classification but show larger variability. Data-level methods become more competitive in multi-class settings, especially when evaluated by G-mean. Overall, the results suggest that method selection should be family- and task-aware rather than based on a single global ranking.
\end{tcolorbox}

\subsection{Performance across Dataset Characteristics}

We next examine how method effectiveness changes with dataset characteristics. Datasets are grouped according to sample size, feature dimensionality, imbalance severity, and the presence of missing values. Datasets containing missing values are excluded from the first three analyses so that the observed differences are not confounded by missing-data handling.

\subsubsection{Sample Size Effects}

\begin{figure}[!ht]
    \centering
    \includegraphics[width=0.95\linewidth]{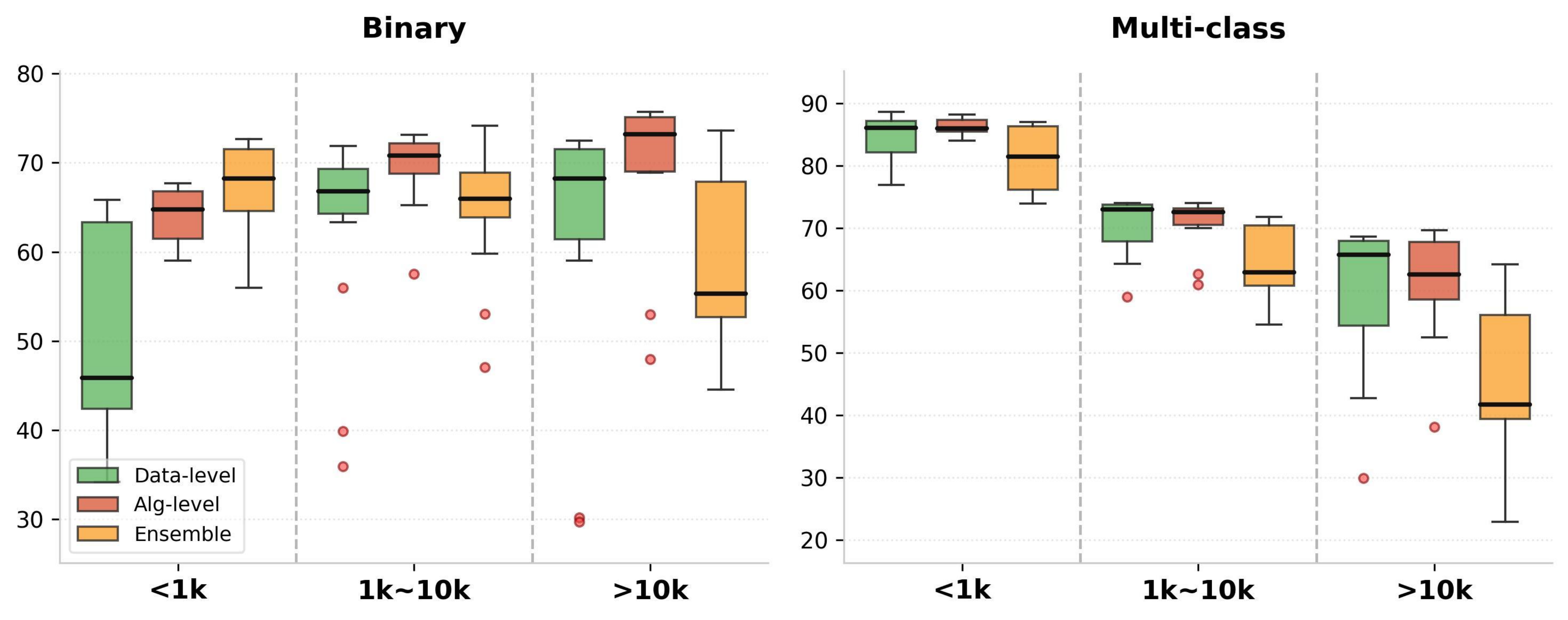}
    \caption{Family-level F1-scores across sample size regimes. Each box shows the distribution of method performance within a family for each dataset size category. Results are reported separately for binary and multi-class tasks.}
    \label{size_f1}
\end{figure}

We group datasets into three regimes: less than 1k, 1k--10k, and greater than 10k samples. Fig.~\ref{size_f1} presents the family-level distributions, while Table~\ref{SS} reports the F1-scores of the top five individual methods in each regime. The corresponding G-mean results are provided in Fig.~\ref{size_g} and Table~\ref{SS_g} in \ref{a.sup}.

For binary classification, Fig.~\ref{size_f1} shows that ensemble-based methods perform best on small datasets but exhibit greater variability. As sample size increases, algorithm-level methods become more competitive and dominate on large datasets, while data-level methods also improve relative to ensemble-based methods. Table~\ref{SS} confirms this trend. Ensemble methods such as SelfPacedEnsemble and BalanceCascadeEnsemble rank highly on small datasets, whereas XGBoost-based algorithm-level methods occupy most top positions when the sample size exceeds 10k. This suggests that the advantage of ensemble methods diminishes with increasing sample size, while algorithm-level methods benefit more from larger datasets.

For multi-class tasks, a different pattern is observed. Fig.~\ref{size_f1} shows that data-level and algorithm-level methods achieve comparable performance across all sample size regimes, both outperforming ensemble-based methods. 
Consistent trends are observed in Table~\ref{SS}, where XGBoostCost and SMOTE-based methods frequently rank among the top-performing approaches across all regimes. Unlike the binary setting, data-level methods appear regularly among the top ranks, while ensemble-based methods rarely do.

\textit{\textbf{Take-away.}}
\textbf{Sample size has a stronger effect on method preference in binary classification, where the advantage shifts from ensemble-based methods at smaller scales to algorithm-level methods at larger scales. In multi-class classification, data-level and algorithm-level methods remain competitive across different sample sizes.}

\begin{table}[!ht]
\centering
\caption{Top five methods in each sample size regime ranked by F1-score for binary and multi-class tasks.}
\label{SS}
\begin{adjustbox}{width=\linewidth}
\begin{tabular}{c|c|c|c|c|c}
\toprule[2pt]
\textbf{Size} & \textbf{Rank} & \textbf{Method} & \textbf{F1-score} & \textbf{Method} & \textbf{F1-score}\\
\midrule[1.5pt]
&&\multicolumn{2}{c|}{Binary}&\multicolumn{2}{c}{Multi-class}\\
\midrule
\multirow{5}{*}{$<$1k} & 1 & AdaUCost & 72.62 $\pm$ 8.65 & PolyfitSMOTE & 88.59 $\pm$ 2.61\\
& 2 & BalanceCascadeEnsemble & 72.48 $\pm$ 8.61 &  XGBoostCost & 88.16 $\pm$ 2.96\\
& 3 & SelfPacedEnsemble & 71.92 $\pm$ 9.23 & SMOTEIPF & 87.92 $\pm$ 2.80\\
& 4 & SMOTEBoost & 71.49 $\pm$ 10.55 & SMOTE & 87.76 $\pm$ 2.78\\
& 5 & SMOTEBagging & 70.18 $\pm$ 9.68 & Lee & 87.57 $\pm$ 3.16\\
\midrule
\multirow{5}{*}{1k--10k} & 1 & SelfPacedEnsemble & 74.13 $\pm$ 3.56 & XGBoostCost & 74.05 $\pm$ 1.61 \\
& 2 & BalanceCascadeEnsemble & 73.76 $\pm$ 3.47 & SMOTE & 74.00 $\pm$ 1.38\\
& 3 & XGBoostWCE & 73.14 $\pm$ 4.74 & SMOTEIPF & 73.93 $\pm$ 1.43\\
& 4 & XGBoostAWE & 72.67 $\pm$ 4.49 & Lee & 73.91 $\pm$ 1.33\\
& 5 & XGBoostASL & 72.34 $\pm$ 5.58 & BorderlineSMOTE & 73.81 $\pm$ 1.29 \\
\midrule
\multirow{5}{*}{$>$10k} & 1 & XGBoostASL & 75.67 $\pm$ 1.86 & XGBoostCost & 69.65 $\pm$ 0.59\\
& 2 & XGBoostWCE & 75.15 $\pm$ 2.06 & SMOTE & 68.66 $\pm$ 0.45\\
& 3 & XGBoostAWE & 75.11 $\pm$ 1.71 & SMOTETomek & 68.63 $\pm$ 0.39\\
& 4 & XGBoostACE & 74.93 $\pm$ 1.88 & BorderlineSMOTE & 68.54 $\pm$ 0.49\\
& 5 & SelfPacedEnsemble & 73.59 $\pm$ 1.51 & XGBoostFL & 68.08 $\pm$ 0.62\\

\bottomrule[2pt]
\end{tabular}
\end{adjustbox}
\end{table}

\subsubsection{Feature Dimensionality}

\begin{figure}[!ht]
    \centering
    \includegraphics[width=0.95\linewidth]{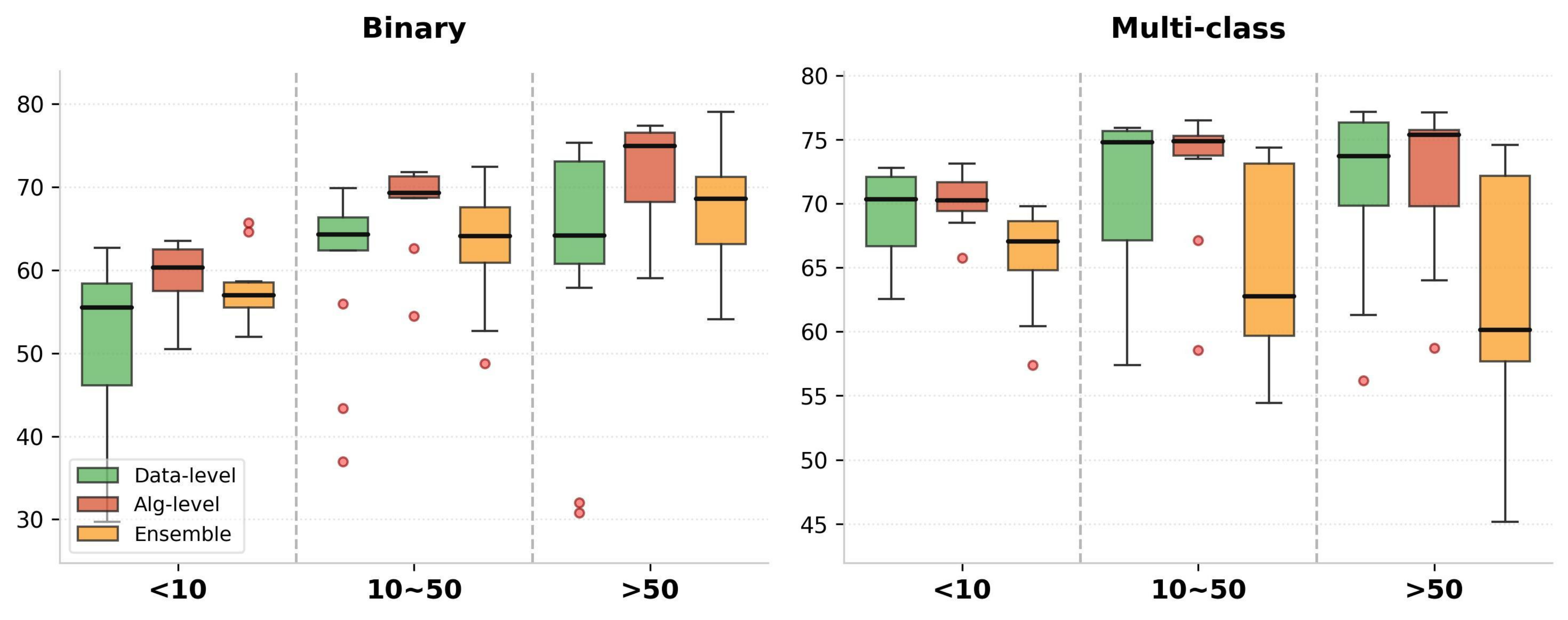}
    \caption{Family-level F1-scores across feature dimensionality regimes. Each box shows the distribution of method performance within a family for each feature group. Results are reported separately for binary and multi-class tasks.}
    \label{dim_f1}
\end{figure}

We group datasets into three feature-dimensionality regimes with fewer than 10, 10--50, and more than 50 features. Fig.~\ref{dim_f1} presents the family-level F1-score distributions, while Table~\ref{FD} reports the top five individual methods and their F1-scores in each regime. The corresponding G-mean results are provided in Fig.~\ref{dim_g} and Table~\ref{FD_g} in \ref{a.sup}.

For binary classification, Fig.~\ref{dim_f1} shows that algorithm-level methods consistently achieve the highest median performance across all feature regimes, with their advantage becoming more pronounced as dimensionality increases. Ensemble methods also perform well and frequently appear among the top-performing methods, although their overall distributions are less stable. In contrast, data-level methods exhibit large variability, including several severe outliers, indicating sensitivity to high-dimensional feature spaces.
The rankings in Table~\ref{FD} further support these observations. Across all feature regimes, the top-performing methods are dominated by ensemble-based and algorithm-level approaches, including SelfPacedEnsemble and multiple XGBoost-based variants. This suggests that increasing dimensionality does not make resampling more advantageous for binary tasks; methods that modify the learning process or aggregate multiple learners remain more reliable.

For multi-class tasks, a different pattern emerges. Fig.~\ref{dim_f1} shows that data-level and algorithm-level methods achieve comparable performance across all feature regimes, both outperforming ensemble-based methods. The performance gap between families increases as dimensionality grows, with ensemble methods showing the largest degradation.
Table~\ref{FD} confirms this trend, with XGBoostCost and SMOTE-based methods consistently appearing among the top-performing approaches. Both data-level and algorithm-level methods maintain strong performance, although each family still contains several underperforming methods. Ensemble methods rarely appear among the top-performing methods and exhibit weaker robustness in high-dimensional settings. This indicates that resampling remains effective for multi-class imbalance even in higher-dimensional feature spaces, while the advantage of ensemble-based strategies does not extend as consistently to these settings.

\textit{\textbf{Take-away.}}
\textbf{Feature dimensionality affects method families differently. Algorithm-level methods remain reliable across dimensionality regimes, data-level methods are more competitive for multi-class than binary tasks, and ensemble-based methods become less favorable as dimensionality increases.}

\begin{table}[!ht]
\centering
\caption{Top five methods in each feature dimensionality regime ranked by F1-score for binary and multi-class tasks.}
\label{FD}
\begin{adjustbox}{width=\linewidth}
\begin{tabular}{c|c|c|c|c|c}
\toprule[2pt]
\textbf{Dimension} & \textbf{Rank} & \textbf{Method} & \textbf{F1-score}& \textbf{Method} & \textbf{F1-score}\\
\midrule[1.5pt]
&&\multicolumn{2}{c|}{Binary}&\multicolumn{2}{c}{Multi-class}\\
\midrule
\multirow{5}{*}{$<$10} & 1 & SelfPacedEnsemble & 65.67 $\pm$ 6.14 & XGBoostCost & 75.38 $\pm$ 2.68\\
& 2 & SMOTEBagging & 64.61 $\pm$ 6.65 & SMOTE & 75.11 $\pm$ 2.28\\
& 3 & XGBoostASL & 63.50 $\pm$ 8.07 & SMOTEIPF & 75.02 $\pm$ 2.33\\
& 4 & XGBoostWCE & 63.46 $\pm$ 7.83 & Tomek & 74.67 $\pm$ 2.83\\
& 5 & Lee & 62.70 $\pm$ 8.13 &  Lee & 74.52 $\pm$ 2.36\\
\midrule
\multirow{5}{*}{10--50} & 1 & SelfPacedEnsemble & 72.41 $\pm$ 3.11 & XGBoostCost & 76.49 $\pm$ 1.35\\
& 2 & BalanceCascadeEnsemble & 71.89 $\pm$ 3.27 & BorderlineSMOTE & 75.90 $\pm$ 0.96\\
& 3 & XGBoostWCE & 71.80 $\pm$ 4.15 & Lee & 75.85 $\pm$ 1.26\\
& 4 & XGBoostAWE & 71.63 $\pm$ 3.27 & SMOBD & 75.69 $\pm$ 1.11\\
& 5 & XGBoostACE & 71.28 $\pm$ 5.40 & SMOTEIPF & 75.68 $\pm$ 1.23\\
\midrule
\multirow{5}{*}{$>$50} & 1 & BalanceCascadeEnsemble & 79.01 $\pm$ 5.46 & SMOTE & 77.15 $\pm$ 0.60\\
& 2 & SelfPacedEnsemble & 78.95 $\pm$ 5.85 & XGBoostCost & 77.11 $\pm$ 0.66\\
& 3 & XGBoostWCE & 77.34 $\pm$ 6.13 & SMOTETomek & 76.94 $\pm$ 0.58\\
& 4 & XGBoostASL & 77.23 $\pm$ 7.28 & BorderlineSMOTE & 76.92 $\pm$ 0.64\\
& 5 & XGBoostAWE & 76.60 $\pm$ 7.05 & SMOBD & 76.43 $\pm$ 0.55\\

\bottomrule[2pt]
\end{tabular}
\end{adjustbox}
\end{table}

\subsubsection{Imbalance Severity}

\begin{figure}[!ht]
    \centering
    \includegraphics[width=0.95\linewidth]{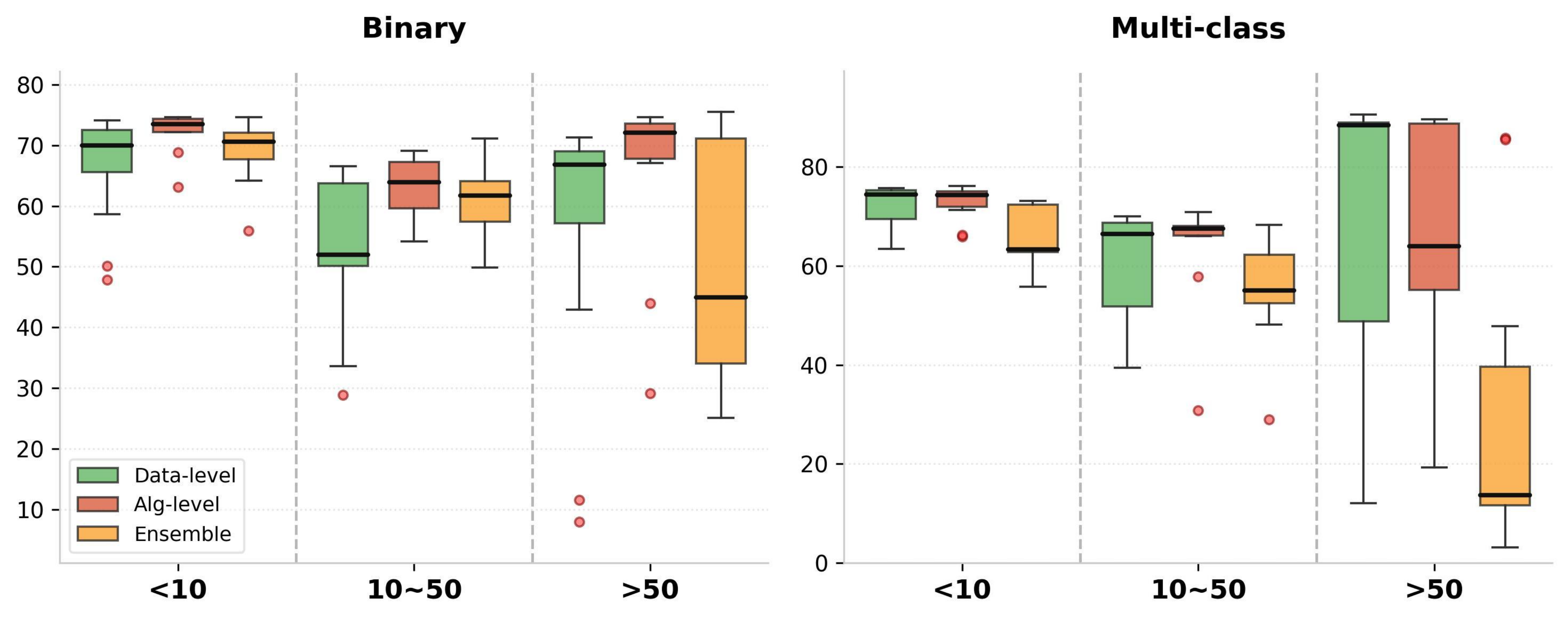}
    \caption{Family-level F1-scores across imbalance severity regimes. Each box shows the distribution of method performance within a family for each imbalance group. Results are reported separately for binary and multi-class tasks.}
    \label{IR_f1}
\end{figure}

We group datasets into three imbalance-ratio levels, less than 10, 10--50, and greater than 50, and analyze performance at the family level in Fig.~\ref{IR_f1}, while Table~\ref{IR} reports the top five individual methods in each regime. Here we report only the F1-score results; the corresponding G-mean results are provided in Fig.~\ref{ir_g} and Table~\ref{IR_g} in \ref{a.sup}.

For binary classification, Fig.~\ref{IR_f1} shows that algorithm-level methods maintain strong performance across all imbalance regimes, while ensemble-based methods remain competitive but exhibit greater variability. Data-level methods are generally less competitive under low and moderate imbalance, but improve under severe imbalance. Table~\ref{IR} further shows that SelfPacedEnsemble ranks first in all three regimes, followed mainly by XGBoost-based algorithm-level methods. This indicates that increasing imbalance severity does not substantially weaken the advantage of algorithm-level and strong ensemble methods in binary tasks.

For multi-class classification, Fig.~\ref{IR_f1} shows that data-level and algorithm-level methods remain comparable across imbalance regimes, whereas ensemble-based methods become less stable as imbalance increases. Table~\ref{IR} confirms that XGBoostCost and SMOTE-based methods dominate the low and moderate regimes, while EditedNearestNeighbours ranks first under severe imbalance. However, the original XGBoost also performs strongly in this regime, suggesting that extreme imbalance does not necessarily make specialized imbalance handling more beneficial.

\textit{\textbf{Take-away.}}
\textbf{Algorithm-level methods remain reliable across imbalance levels, while data-level methods become more competitive as imbalance increases, particularly in multi-class tasks. Ensemble-based methods remain effective for binary classification but are less robust in multi-class settings.}

\begin{table}[!ht]
\renewcommand\arraystretch{1.2}
\centering
\caption{Top five methods in each imbalance severity regime ranked by F1-score for binary and multi-class tasks.}
\label{IR}
\begin{adjustbox}{width=\linewidth}
\begin{tabular}{c|c|c|c|c|c}
\toprule[2pt]
\textbf{Ratio} & \textbf{Rank} & \textbf{Method} & \textbf{F1-score} & \textbf{Method} & \textbf{F1-score}\\
\midrule[1.5pt]
&&\multicolumn{2}{c|}{Binary}&\multicolumn{2}{c}{Multi-class}\\
\midrule
\multirow{5}{*}{$<$10} & 1 & SelfPacedEnsemble & 74.65 $\pm$ 2.68 & XGBoostCost & 76.14 $\pm$ 1.70\\
& 2 & XGBoostASL & 74.61 $\pm$ 2.68 & SMOTE & 75.75 $\pm$ 1.56\\
& 3 & XGBoostAWE & 74.50 $\pm$ 2.53 & Lee & 75.59 $\pm$ 1.62\\
& 4 & XGBoostWCE & 74.43 $\pm$ 3.19 & SMOTETomek & 75.49 $\pm$ 1.56\\
& 5 & BalanceCascadeEnsemble & 74.29 $\pm$ 2.13 & PolyfitSMOTE & 75.39 $\pm$ 1.45\\
\midrule
\multirow{5}{*}{10--50} & 1 & SelfPacedEnsemble & 71.13 $\pm$ 6.31 & XGBoostCost & 70.89 $\pm$ 1.26\\
& 2 & BalanceCascadeEnsemble & 69.59 $\pm$ 8.75 & SMOTE & 70.04 $\pm$ 0.85\\
& 3 & XGBoostASL & 69.11 $\pm$ 9.63 & BorderlineSMOTE & 69.93 $\pm$ 0.68\\
& 4 & XGBoostWCE & 69.11 $\pm$ 8.08 & PolyfitSMOTE & 69.56 $\pm$ 0.54\\
& 5 & SMOTEBagging & 67.28 $\pm$ 7.25 & SMOTEIPF & 68.89 $\pm$ 1.98\\
\midrule
\multirow{5}{*}{$>$50} & 1 & SelfPacedEnsemble & 75.47 $\pm$ 5.13 & EditedNearestNeighbours & 90.63 $\pm$ 0.66\\
& 2 & XGBoostAWE & 74.64 $\pm$ 4.90 & XGBoost & 90.27 $\pm$ 0.62\\
& 3 & XGBoostFL & 74.18 $\pm$ 4.14 & OneSidedSelection & 89.74 $\pm$ 1.29\\
& 4 & XGBoostCBE & 73.61 $\pm$ 5.28 & XGBoostFL & 89.64 $\pm$ 1.39\\
& 5 & XGBoostWCE & 73.48 $\pm$ 5.08 & XGBoostCBE & 89.27 $\pm$ 1.09\\

\bottomrule[2pt]
\end{tabular}
\end{adjustbox}
\end{table}

\subsubsection{Missing Value Compatibility}
We evaluate the native compatibility of imbalanced learning methods with missing values. Table~\ref{T.f1miss} reports the F1-score results for compatible methods, with the corresponding G-mean results provided in Table~\ref{T.gmiss} in \ref{a.sup}.

Data-level methods generally cannot directly handle missing values because many rely on distance computation, neighborhood search, or interpolation. For example, SMOTE and its variants require complete observations to generate synthetic samples. Consequently, these methods typically require imputation before resampling, introducing an additional preprocessing step.

Algorithm-level methods provide better native compatibility. In particular, XGBoost-based methods benefit from sparsity-aware split finding, allowing them to process incomplete data without explicit imputation. As shown in Table~\ref{T.f1miss}, these methods achieve strong and stable performance in both binary and multi-class settings. DecisionTreeCost and RandomForestCost are also applicable but generally achieve lower performance. 

Ensemble-based methods exhibit mixed behavior. Some methods are compatible with missing values because their base learners, such as decision-tree-based models, can naturally process incomplete data. As shown in Table~\ref{T.f1miss}, SelfPacedEnsemble and BalanceCascadeEnsemble achieve competitive performance, particularly in binary tasks. However, other ensemble methods show more moderate performance, and their compatibility strongly depends on the underlying base learner and training framework.

\textit{\textbf{Take-away.}}
\textbf{Algorithm-level methods provide the most practical solution when missing values are present because they combine native compatibility with competitive performance. Data-level methods generally require additional imputation, while the applicability of ensemble methods depends on their base learners and resampling procedures.}

\begin{table}[!ht]
\centering
\caption{Performance of methods that support missing values. We report F1-scores for both binary and multi-class tasks. Methods without native support are excluded.}
\label{T.f1miss}
\begin{tabular}{c|c|c}
\toprule[2pt]
\textbf{Method} & \textbf{Binary} & \textbf{Multi-class}\\
\midrule[1.5pt]
\multicolumn{3}{c}{Algorithm-level}\\
\midrule
XGBoostASL & 72.62 $\pm$ 5.48 & 72.55 $\pm$ 3.25\\
XGBoostACE & 72.07 $\pm$ 6.16 & 72.65 $\pm$ 2.76\\
XGBoostAWE& 72.54 $\pm$ 4.57 & 71.66 $\pm$ 3.15\\
XGBoostFL & 70.14 $\pm$ 5.84 & 74.69 $\pm$ 1.76\\
XGBoostWCE & 72.96 $\pm$ 4.91 & 74.61 $\pm$ 1.31\\
XGBoostCBE & 69.58 $\pm$ 5.14 & 74.69 $\pm$ 1.64\\
XGBoostCost & 69.90 $\pm$ 4.90 & 75.83 $\pm$ 1.57\\
DecisionTreeCost & 61.35 $\pm$ 4.82 & 64.90 $\pm$ 2.04\\
RandomForestCost & 67.83 $\pm$ 4.83 & 70.65 $\pm$ 1.64\\
\midrule[1.5pt]
\multicolumn{3}{c}{Ensemble}\\
\midrule
SelfPacedEnsemble & 73.88 $\pm$ 4.01 & 70.45 $\pm$ 1.45\\
BalanceCascadeEnsemble & 72.51 $\pm$ 4.55 & 69.28 $\pm$ 1.47\\
UnderBagging & 61.55 $\pm$ 3.53 & 71.10 $\pm$ 1.13\\
OverBagging & 67.86 $\pm$ 4.70 & 72.14 $\pm$ 1.32\\
\bottomrule[2pt]
\end{tabular}
\end{table}

\begin{tcolorbox}[colback=gray!10, colframe=black!50, title=\textbf{Summary for RQ2: Performance Across Dataset Characteristics}]
The effectiveness of imbalanced learning methods is strongly dependent on dataset characteristics. Algorithm-level methods demonstrate consistently strong and stable performance across different regimes, particularly in the presence of missing values. In contrast, data-level methods become more competitive in multi-class and highly imbalanced scenarios. Ensemble-based methods perform well on small datasets but exhibit reduced robustness as sample size, feature dimensionality, or imbalance severity increases. These results further demonstrate that no single method is universally optimal, and that method effectiveness is highly regime-dependent.
\end{tcolorbox}

\subsection{Scalability and Computational Efficiency}

Predictive performance alone is insufficient for method comparison when datasets become large or complex. We therefore analyze training-time scalability with respect to sample size, feature dimensionality, and number of classes.

We use \texttt{scikit-learn}~\cite{pedregosa2011scikit} to generate synthetic datasets. Each experiment varies one data characteristic while holding the remaining characteristics fixed. Unless otherwise specified, the baseline configuration contains 10000 samples, 10 features, two classes, and an imbalance ratio of 1:9. All methods are evaluated in the same CPU environment. Figures~\ref{scal1}--\ref{scal3} show method-level training times, while detailed family-level statistics are provided in Tables~\ref{scal1time}--\ref{scal3time} in \ref{a.sup}. 
For readability in figures, abbreviations of all evaluated methods are provided in Table~\ref{T.abbr} in \ref{a.imp}.

\subsubsection{Scalability across Sample Sizes}

\begin{figure}[!ht]
    \centering
    \includegraphics[width=0.9\linewidth]{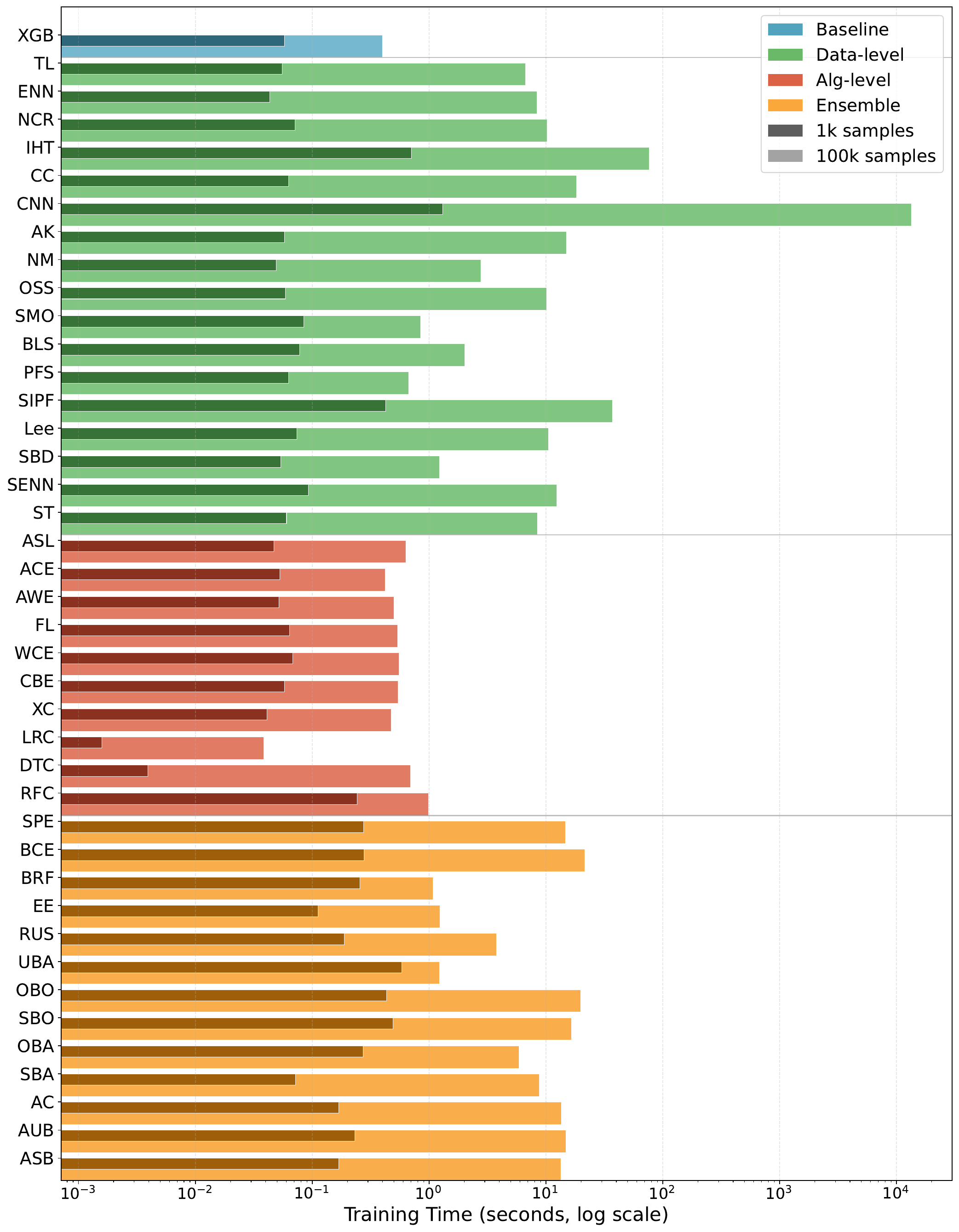}
    \caption{Training time ($\log_{10}(\text{seconds})$) of different methods under increasing sample sizes. Each method is evaluated on datasets with 1k and 100k samples. Bar length indicates training time on a logarithmic scale. Colors represent method families, and color intensity indicates dataset size.}
    \label{scal1}
\end{figure}

We measure training time on datasets containing 1k and 100k samples, as shown in Fig.~\ref{scal1}.

Algorithm-level methods exhibit the most consistent scaling behavior. Most XGBoost-based variants remain close to the baseline and show only moderate increases in training time. This suggests that incorporating imbalance awareness through the objective or sample weighting introduces relatively limited additional cost.

Data-level methods show substantially greater variability. Their computational cost is modest on small datasets but can increase sharply as sample size grows. Methods based on nearest-neighbor search or repeated sample evaluation are particularly expensive. SMOTE-based approaches require neighborhood construction among minority samples, while InstanceHardnessThreshold and CondensedNearestNeighbor repeatedly invoke auxiliary learning or sample-selection procedures.

Ensemble-based methods also incur increasing cost because they repeatedly train base learners on resampled or reweighted datasets. Their growth is generally more predictable than that of the most expensive data-level methods, but their absolute training time remains higher than that of most algorithm-level methods.

\textit{\textbf{Take-away.}}
\textbf{Algorithm-level methods scale most consistently with sample size, whereas neighborhood-based resampling and repeated ensemble training become increasingly expensive on large datasets.}

\subsubsection{Scalability across Feature Dimensions}

\begin{figure}[!ht]
    \centering
    \includegraphics[width=0.9\linewidth]{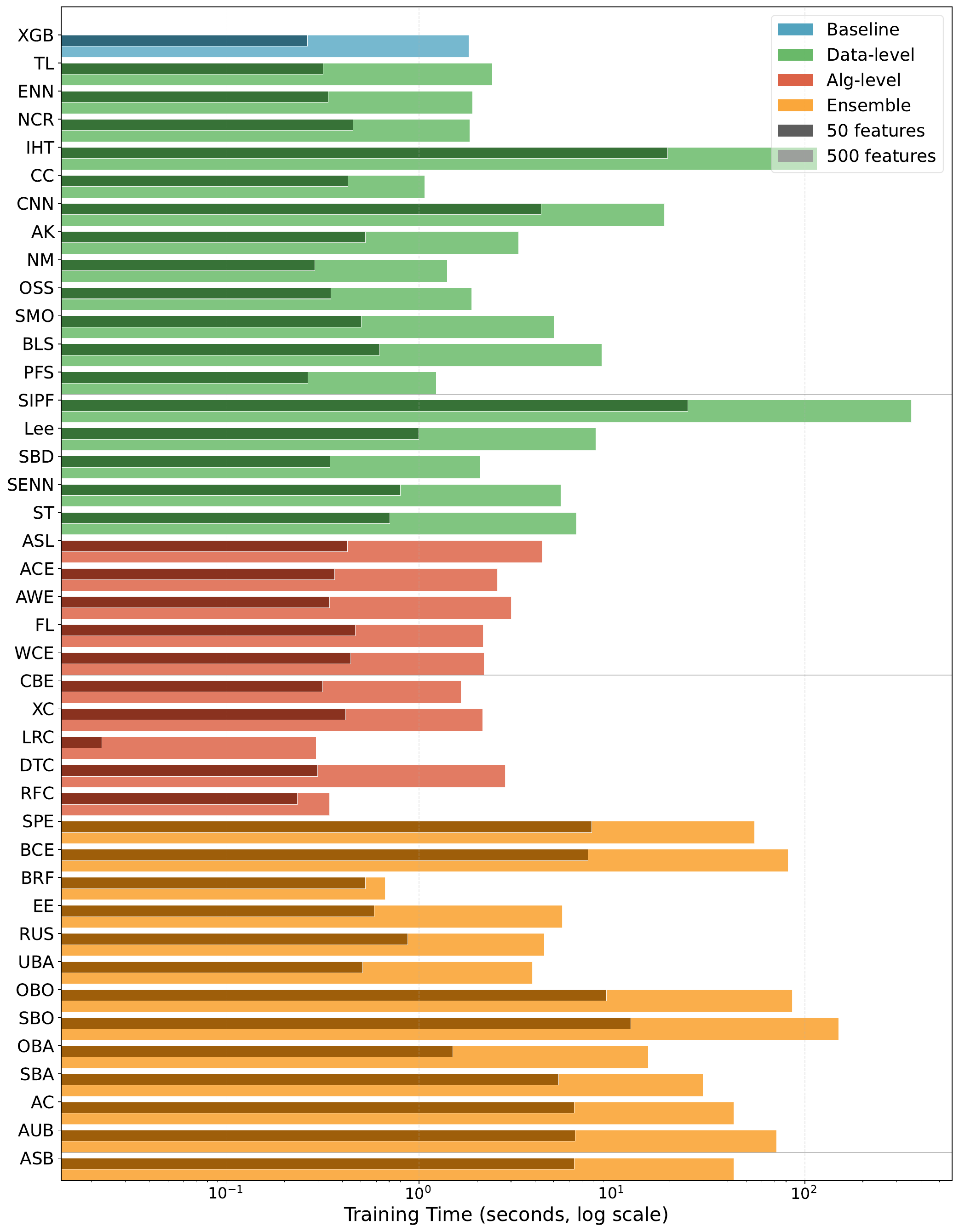}
    \caption{Training time ($\log_{10}(\text{seconds})$) of different methods under increasing feature dimensionality. Each method is evaluated on datasets with 50 and 500 features. Bar length represents training time on a logarithmic scale. Colors represent method families, and color intensity indicates feature dimensionality.}
    \label{scal2}
\end{figure}

We increase the number of features from 50 to 500 and report the results in Fig.~\ref{scal2}.

Most algorithm-level methods exhibit moderate growth and remain close to the XGBoost baseline. LogRegCost is an exception, showing a more noticeable increase because its optimization cost depends directly on the full feature representation.

Data-level methods are more sensitive to dimensionality, especially when they rely on distance computation. InstanceHardnessThreshold, CondensedNearestNeighbor, and SMOTEIPF become substantially more expensive in the high-dimensional regime. Other methods such as TomekLinks and NearMiss remain comparatively efficient, showing that computational behavior can vary considerably within the same family.

Ensemble-based methods incur the largest aggregate overhead because both model fitting and, in many cases, resampling must be repeated. Nevertheless, BalancedRandomForest and EasyEnsemble remain relatively efficient compared with other ensemble designs.

\textit{\textbf{Take-away.}}
\textbf{Feature dimensionality has the strongest computational effect on distance-based data-level methods and repeated ensemble procedures, while most algorithm-level methods retain stable efficiency.}

\subsubsection{Scalability across Class Numbers}

\begin{figure}[!ht]
    \centering
    \includegraphics[width=0.9\linewidth]{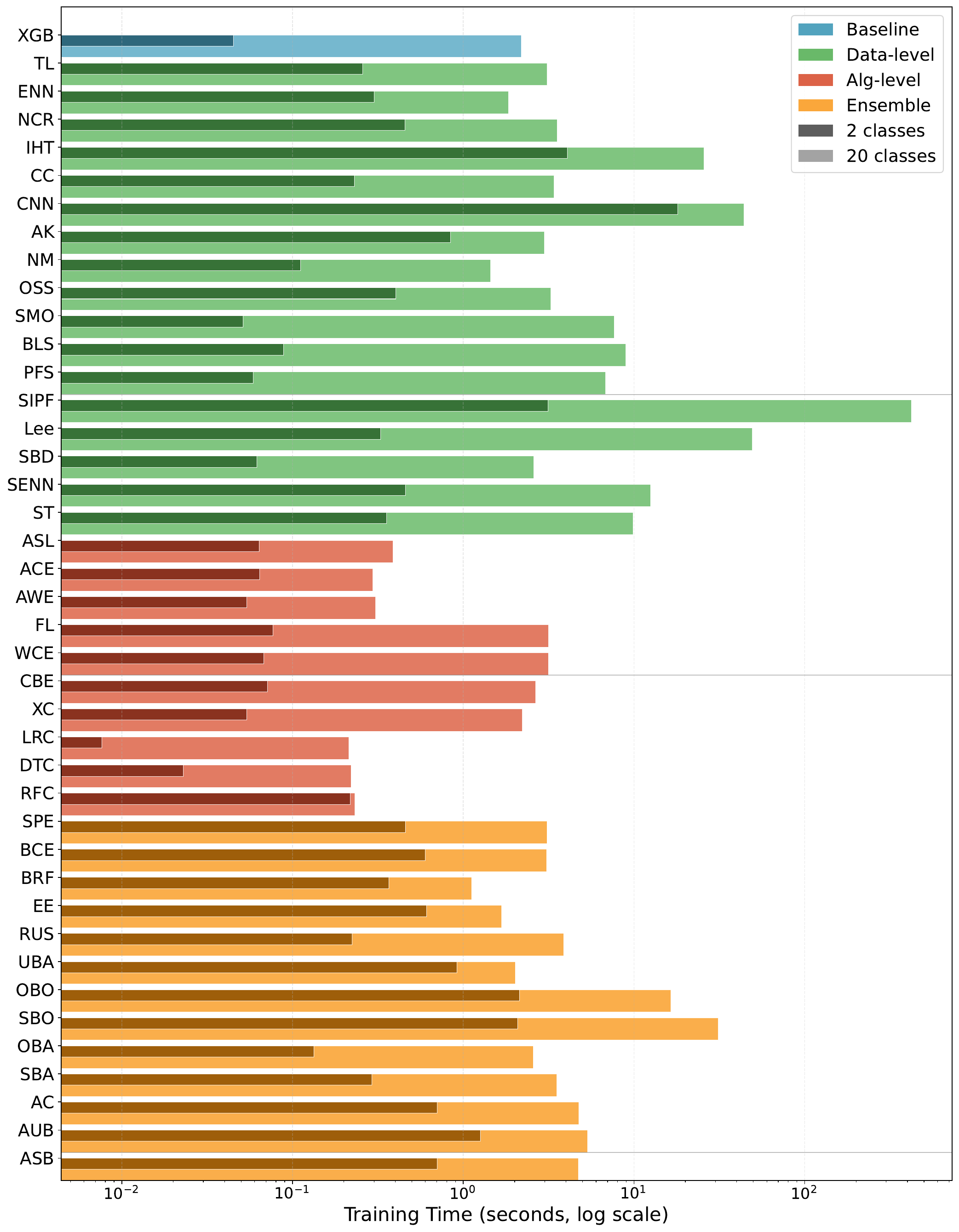}
    \caption{Training time ($\log_{10}(\text{seconds})$) of different methods under increasing class numbers. Each method is evaluated on datasets with 2 and 20 classes. Bar length represents training time on a logarithmic scale. Colors indicate method families, while color intensity represents the number of classes.}
    \label{scal3}
\end{figure}

We compare datasets with 2 and 20 classes, as shown in Fig.~\ref{scal3}.

Algorithm-level methods remain among the most computationally efficient approaches. Most XGBoost-based variants exhibit only limited increases in training time as the number of classes grows, remaining close to the baseline across both regimes. Simpler methods such as LogRegCost and DecisionTreeCost are particularly efficient, although their predictive performance is generally weaker than that of XGBoost-based methods.

Ensemble methods exhibit mixed scalability behavior. Methods such as SelfPacedEnsemble, BalanceCascadeEnsemble, BalancedRandomForest, and EasyEnsemble maintain relatively stable training time as the number of classes increases. In contrast, several bagging-based oversampling methods remain computationally expensive across both class regimes, resulting in substantially larger overhead than other families.

Data-level methods show the largest variability. Simpler undersampling methods such as EditedNearestNeighbors and NearMiss remain highly efficient, while methods involving iterative classifier-based procedures during the resampling process tend to incur substantially higher computational costs. In particular, InstanceHardnessThreshold, CondensedNearestNeighbor, and SMOTE-IPF consistently exhibit high computational cost across both binary and multi-class settings.

\textit{\textbf{Take-away.}}
\textbf{Class number has a smaller and less systematic effect on runtime than sample size or dimensionality. Algorithm-level methods remain the most stable family, whereas the cost of data-level and ensemble methods depends strongly on their internal design.}

\begin{tcolorbox}[
    colback=gray!10,
    colframe=black!50,
    title=\textbf{Summary for RQ3: Scalability and Computational Efficiency}
]
Algorithm-level methods exhibit the most stable computational behavior as sample size, feature dimensionality, and class number increase. Data-level methods may become expensive when they rely on neighborhood search, iterative cleaning, or auxiliary classifiers, while ensemble-based methods incur repeated training overhead. Scalability therefore depends not only on the broad method family but also on the specific operations used within each algorithm.
\end{tcolorbox}

\subsection{Practical Recommendations}

\begin{figure}[!ht]
    \centering
    \includegraphics[width=0.95\textwidth]{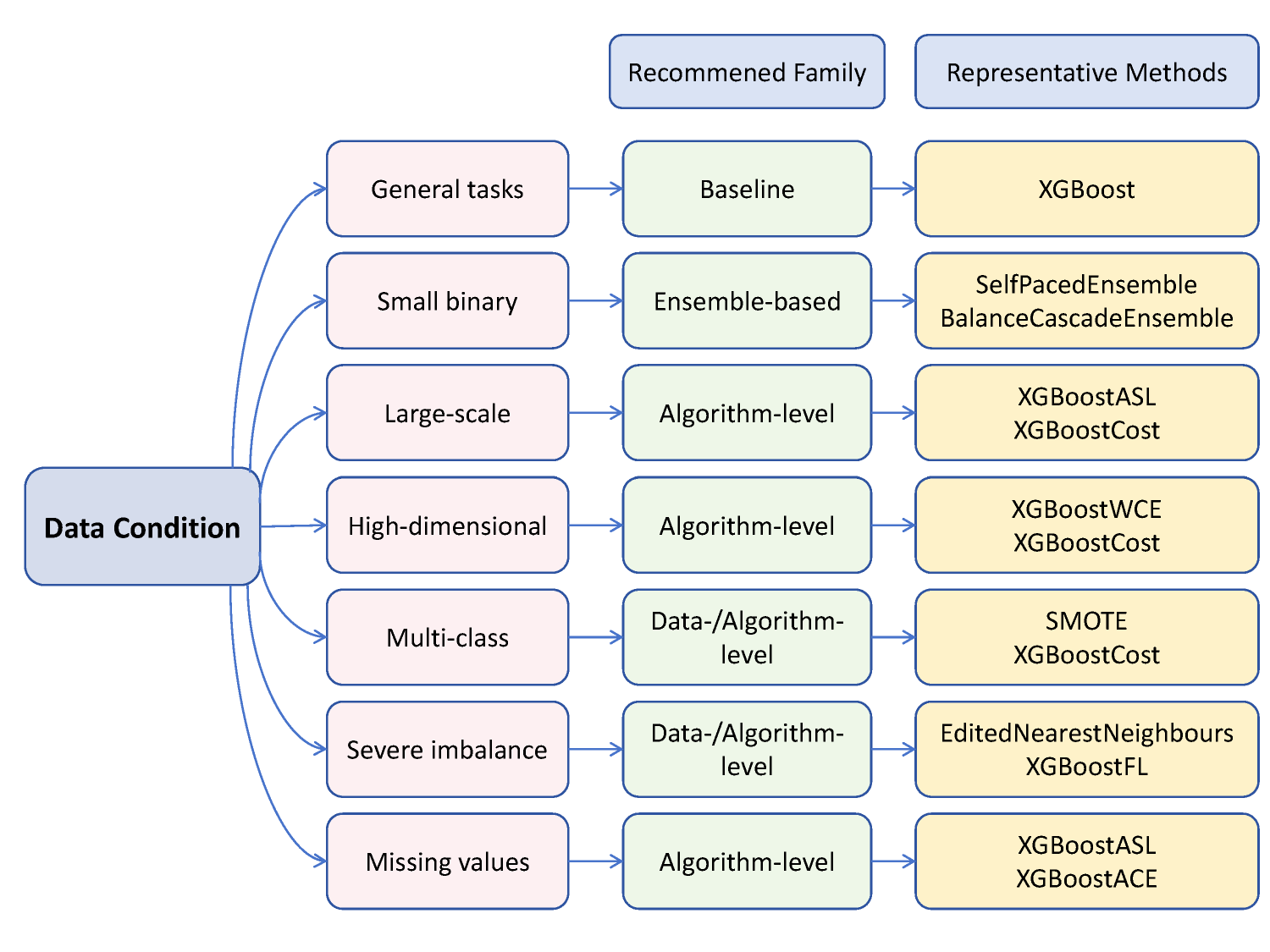}
    \caption{Regime-aware recommendations for selecting imbalanced learning methods under different tabular data conditions.}
    \label{f.recom}
\end{figure}

The empirical findings support a regime-aware approach to method selection. Rather than treating imbalance handling as an isolated preprocessing decision, practitioners should jointly consider the task structure, dataset characteristics, baseline learner, and available computational resources. Fig.~\ref{f.recom} summarizes the main recommendations.

A strong tabular learner should first be established as a reference. XGBoost remains competitive across many datasets, particularly in multi-class classification, and therefore provides a useful baseline against which the benefit of an imbalance-specific technique can be assessed.

For small binary datasets, ensemble methods such as SelfPacedEnsemble and BalanceCascadeEnsemble are strong candidates because aggregation can compensate for limited minority observations. However, their additional training cost should be considered. When dataset size or dimensionality increases, XGBoost-based algorithm-level methods generally provide a more reliable combination of predictive performance and computational efficiency.

For multi-class tasks, resampling methods become more competitive. SMOTE and its variants are strong candidates when imbalance occurs across multiple classes, while XGBoostCost provides a stable and relatively simple alternative. Under severe imbalance, structured undersampling methods such as EditedNearestNeighbours and OneSidedSelection can also be effective.

For datasets with missing values, methods with native incomplete-data support should be preferred. XGBoost-based algorithm-level approaches can handle missing values without inserting a separate imputation-resampling pipeline. Most data-level methods require imputation before neighborhood computation or interpolation, which may introduce additional bias or uncertainty.

In summary, method selection should jointly consider task type, dataset regime, base learner, and computational budget. Ensemble methods are most attractive for small binary datasets, data-level methods are particularly competitive in multi-class and severely imbalanced settings, and algorithm-level methods provide the most consistently reliable option across scalability and missing-value conditions.

\section{Open Challenges and Future Directions}
\label{s:future}

The survey and empirical results reveal several unresolved challenges in tabular imbalanced learning. These challenges concern not only the design of more accurate algorithms, but also their adaptability, scalability, evaluation, and deployment across heterogeneous data regimes.

\paragraph{\textbf{From Static Methods to Regime-Aware Adaptation}}
Most existing methods employ a fixed imbalance-handling strategy regardless of dataset characteristics. For example, a resampling ratio, class-weighting rule, or ensemble configuration is usually chosen before training and then applied uniformly. Our results show, however, that the relative effectiveness of these choices changes with sample size, feature dimensionality, imbalance severity, missingness, and task type.

An important direction is therefore to develop methods that adapt their imbalance-handling mechanism to the observed data regime. Such methods could automatically determine whether resampling, objective modification, or model aggregation is appropriate and adjust their strength during training. This would move the field from manually selected algorithms toward data-dependent and adaptive imbalance handling.

\paragraph{\textbf{Missing Values and Imbalance Should Be Modeled Jointly}}
Missingness and class imbalance are often treated as separate preprocessing problems. In practice, however, missing values may occur disproportionately across classes or may themselves contain predictive information. Applying generic imputation before imbalance handling can distort minority distributions, while synthetic resampling after imputation may amplify imputation artifacts.

Future methods should jointly model missingness and imbalance rather than processing them sequentially. Promising directions include class-conditional imputation, missingness-aware resampling, objectives that account for uncertainty in imputed minority samples, and tree-based or generative methods that can handle both phenomena natively.

\paragraph{\textbf{Scalable Resampling and Ensemble Design}}
The scalability analysis shows that several data-level and ensemble methods become expensive on large or high-dimensional datasets. The main bottlenecks include exact nearest-neighbor search, repeated classifier fitting, iterative data cleaning, and large ensemble sizes.

Future research should investigate approximate nearest-neighbor search, prototype-based resampling, mini-batch resampling, distributed implementation, and adaptive ensemble pruning. A key objective is to retain minority-class coverage and model diversity without allowing computational cost to grow excessively with sample size or dimensionality.

\paragraph{\textbf{Multi-Class and Multi-Minority Imbalance}}
Many classical imbalance-handling methods were originally designed for binary classification. Multi-class imbalance introduces additional difficulties because several minority classes may have different sample sizes, overlap structures, and pairwise relationships. A method that improves one minority class may simultaneously degrade another.

Future work should move beyond one-versus-rest adaptations and explicitly model relations among multiple minority classes. Class-specific resampling, pairwise cost modeling, hierarchical learning, and adaptive objectives may provide more principled solutions.

\paragraph{\textbf{Deep Learning for Tabular Imbalance}}
Recent deep learning \cite{borisov2022deep,jiang2026representation} and foundation models \cite{hollmann2022tabpfn,hollmann2025accurate} for tabular data introduce new opportunities and challenges. Their learned representations may reduce sensitivity to sparse minority samples, but large pretrained models may also inherit majority-class bias from their pretraining data. Moreover, the interaction between in-context learning \cite{qu2025tabicl,zhang2025limix}, parameter-efficient adaptation \cite{houlsby2019parameter,luo2026tablora,hu2022lora}, and imbalance-aware objectives \cite{wang2020imbalance, luo2025robust, tian2023synergetic, mushava2024flexible} remains underexplored.

Future work should investigate whether pretrained representations reduce the need for explicit resampling, how imbalance-aware adaptation should be performed, and whether synthetic minority examples can be generated safely in learned representation spaces. Computational cost and accessibility will remain important considerations.

\section{Conclusions}
\label{s:conclu}

This paper has presented a systematic study of tabular imbalanced learning through a unified survey and the \textbf{TILBench} benchmark. By examining a broad range of methods across heterogeneous datasets and computational settings, our study connects differences in methodological design with their practical performance.

The results show that there is no universally best imbalance-handling strategy. Algorithm-level methods generally offer strong and stable performance, particularly when scalability or native missing-value handling is important. Data-level approaches become more competitive in multi-class and severely imbalanced settings, although their effectiveness varies considerably across algorithms. Ensemble-based methods can perform well on small binary datasets, but their variability and repeated training overhead make them less attractive as data scale and complexity increase. Notably, XGBoost alone remains highly competitive in many cases, indicating that additional imbalance handling may provide limited benefits when the base learner is already strong.

Overall, effective method selection should depend on the data regime rather than a single global ranking. A strong tabular learner provides a useful starting point, after which specialized techniques can be considered according to data characteristics and computational requirements. The study also reveals several unresolved challenges and opportunities for further investigation. We hope these findings provide a useful reference for both practical method selection and future research in tabular imbalanced learning.


\section*{Declaration of generative AI use}
During the preparation of this work the authors used ChatGPT in order to improve the language and readability. After using this tool/service, the authors reviewed and edited the content as needed and take full responsibility for the content of the publication.


\appendix
\section{Implementation Details}
\label{a.imp}

Table~\ref{T.data} summarizes the datasets used in this study. Table~\ref{T.hyperparam} presents the hyperparameter search ranges for all methods. Methods using XGBoost as the base learner share the same hyperparameter settings as the baseline model; therefore, only additional method-specific hyperparameters are reported to avoid redundancy. We also provide the corresponding open-source implementation links in Table~\ref{T.python}.

\begin{table}[!ht]
\footnotesize
\centering
\caption{Dataset description. \#S means the sample number; \#F indicates the feature number; \#C gives the number of class; \#M is the number of samples with missing values; IR is the imbalanced ratio. For binary and multi-class classification, IR is the ratio of the most frequent class to the least frequent class. Dataset name with an asterisk (*) is an abbreviation. Full names are: (1) o-s-i: online-shoppers-intention (2) I-A: Internet-Advertisements (3) P-D-H: Pulsar-Dataset-HTRU2 (4) C-C-F-D: CreditCardFraudDetection (5) G-P-S-P: GesturePhaseSegmentationProcessed (6) j-c-2-r-e-c: jungle\_chess\_2pcs\_raw\_endgame\_complete.}
\label{T.data}
\begin{adjustbox}{width=1.0\textwidth}
\begin{tabular}{lccccc|lccccc}
\toprule[1.5pt]
Name & \#S & \#F & \#C & \#M & IR & Name & \#S & \#F & \#C & \#M & IR\\
\midrule[1pt]

\multicolumn{12}{c}{Binary} \\
\hline
breast-w & 699 & 9 & 2 &16 & 1.90 & bank32nh & 8192 & 32 & 2 & 0 & 2.22 \\
ada& 4147 & 48 & 2 & 0 & 3.03 & adult & 48842 & 14 & 2 & 0 & 3.18\\
kc2 & 522 & 21 & 2 & 0 & 3.88 & o-s-i* & 12330 & 17 & 2 & 0 & 5.46\\
churn & 5000 & 20 & 2 & 0 & 6.07 & I-A* & 3279 & 1558 & 2 & 0 & 6.14\\
kick & 72983 & 32 & 2 & 383 & 7.13 & pc4 & 1458 & 37 & 2 & 0 & 7.19\\
bank-marketing& 45211 & 16 & 2 & 0 & 7.55 & ecoli & 336 & 7 & 2 & 0& 8.60\\
page-blocks & 5473 & 10 & 2 & 0 & 8.77 & satimage & 6435 & 36 & 2 & 0 & 9.28\\
sick\_euthyroid & 3163 & 42 & 2 & 0 & 9.80 & P-D-H* & 17898 & 8 & 2 & 0 & 9.92\\
vowel & 990 & 12 & 2 & 0 & 10.00 & spectrometer & 531 & 93 & 2 & 0 & 10.80\\
car\_eval\_34 & 1728 & 21 & 2 & 0 & 11.90 & isolet & 7797 & 617 & 2 & 0 & 12.00\\
us\_crime & 1994 & 100 & 2 & 0 & 12.29 & libras\_move & 360 & 90 & 2 & 0 & 14.00\\
sick & 3772 & 29 & 2 & 3772 &15.33 & thyroid\_sick & 3772 & 52 & 2 & 0 &15.33\\
arrhythmia & 452 & 278& 2 & 0 & 17.08 & oil & 937 & 49 & 2 & 0 & 21.85\\
yeast\_me2 & 1484 & 8 & 2 & 0 & 28.09 & webpage & 34780 & 300 & 2 &0 & 34.45\\
mammography& 11183 & 6 & 2 & 0 & 42.01 & APSFailure& 76000 & 170 & 2 & 75244 & 54.27\\
dis & 3772 & 29 & 2 & 0 & 64.03 & Satellite& 5100 & 36 & 2 & 0 & 67.00\\
protein\_homo & 145751 & 74 & 2 & 0 & 111.46 & C-C-F-D* & 284807 & 30 & 2 & 0 & 577.88\\
\midrule[1pt]

\multicolumn{12}{c}{Multi-class} \\
\hline
MiceProtein & 1080 & 77 & 8 & 528 & 1.43 & wine & 178 & 13 & 3 & 0 & 1.48\\
contraceptive & 1473 & 9 & 3 & 0 & 1.89 & cmc & 1473 & 9 & 3 & 0 & 1.89\\
eucalyptus & 736 & 19 & 5 & 95 & 2.04 & hayes-roth & 160 & 4 & 3& 0 & 2.10\\
splice & 3190 & 60 & 3 & 0 & 2.16 & dna & 3186 & 180 & 3 & 0 & 2.16\\
satimage & 6430 & 36 & 6 & 0 &2.45 & G-P-S-P* & 9873 & 32 & 5 & 0 & 2.96\\
fabert & 8237 & 800 & 7 & 0 & 3.84 & Diabetes130US & 101766 & 49 & 3 & 0 & 4.83\\
newthyroid & 215 & 5 & 3 &0& 5.00 & j-c-2-r-e-c* & 44819 & 6 & 3 & 0 & 5.32\\
dermatology & 358 & 34 & 6 & 0 & 5.55 & balance&  625 & 4 & 3 & 0 & 5.88\\
connect-4 & 67557 & 42 & 3 & 0 & 6.90 & okcupid-stem & 50789 & 19 & 3& 1 & 7.45\\
flare & 1066 & 19 & 6 & 0 & 7.70 & volkert & 58310 & 180 & 10 & 0 & 9.41\\
steel-plates-fault & 1941 & 27 & 7 & 0 & 12.24 & jannis & 83733 & 54 & 4 & 0 & 22.83 \\
kropt & 28056 & 6 & 18 & 0 & 168.63 & & & & & & \\

\bottomrule[1.5pt]
\end{tabular}
\end{adjustbox}\\[5mm]
\end{table}

\begin{table}[!ht]
\footnotesize
\centering
\caption{The hyperparameters involved in training are given. Methods in the data-level family, as well as those involving XGBoost in the algorithm-level family, share the same four hyperparameters as the base model, thus we leave out the repeated parts.}
\label{T.hyperparam}
\begin{adjustbox}{width=\textwidth}
\begin{tabular}{lllcl}

\toprule[2pt]
\textbf{Category} & \textbf{Algorithm} & \textbf{Hyperparameter} & \textbf{Type} & \textbf{Range/Values} \\
\midrule[1.5pt]

\multirow{4}{3cm}{Base Model} & \multirow{4}{*}{XGBoost} & max\_depth& int& $[2, 10]$\\
& & alpha & float & $[1e-8, 0.1]$\\
& & lambda & float & $[0.5, 2.0]$\\
& & eta & float & $[0.05, 0.3]$\\
\midrule[1.5pt]

\multirow{17}{2cm}{Data-level Methods} 
& TomekLinks & sampling\_strategy & - & auto \\
\cline{2-5}
& EditedNearestNeighbours & n\_neighbors & int & $[2, 8]$\\
\cline{2-5}
& NeighbourhoodCleaningRule & n\_neighbors & int& $[2, 8]$\\
\cline{2-5}
& InstanceHardnessThreshold &sampling\_strategy & - & auto \\
\cline{2-5}
& ClusterCentroids  &sampling\_strategy & - & auto \\
\cline{2-5}
& CondensedNearestNeighbour & n\_neighbors & int & $[2, 8]$\\
\cline{2-5}
& AllKNN & n\_neighbors & int & $[2, 8]$\\
\cline{2-5}
& NearMiss & n\_neighbors & int & $[2, 8]$\\
\cline{2-5}
& OneSidedSelection & n\_neighbors & int & $[2, 8]$\\
\cline{2-5}
& SMOTE & k\_neighbors & int & $[3, 10]$\\
\cline{2-5}
& BorderlineSMOTE & k\_neighbors & int & $[3, 10]$\\
\cline{2-5}
& PolyfitSMOTE & order & int & $[1, 3]$\\
\cline{2-5}
& SMOTEIPF & n\_neighbors & int & $[3, 10]$\\
\cline{2-5}
& Lee & n\_neighbors & int & $[3, 10]$\\
\cline{2-5}
& SMOBD & eta1 & float & $[0.4, 1.0]$\\
\cline{2-5}
& SMOTEENN & k\_neighbors & int & $[3, 10]$\\
\cline{2-5}
& SMOTETomek & k\_neighbors & int & $[3, 10]$\\

\midrule[1.5pt]

\multirow{18}{3cm}{Algorithm-level Methods} 
& \multirow{3}{*}{XGBoostASL} & r1 & categorical & $[0.0, 0.1]$\\
& & r2 & categorical & $[0.5, 1.0, 2.0]$\\
& & m & float & $[0.05, 0.2]$\\
\cline{2-5}
& XGBoostACE & m & float & $[0.05, 0.2]$\\
\cline{2-5}
& \multirow{2}{*}{XGBoostAWE} & r1 & categorical & $[2.0, 3.0, 5.0]$\\
& & m & float & $[0.05, 0.2]$\\
\cline{2-5}
& XGBoostFL & r1 & categorical & $[0.5, 1.0, 2.0]$\\
\cline{2-5}
& XGBoostWCE & r1 & categorical & $[2.0, 3.0, 5.0]$\\
\cline{2-5}
& XGBoostCBE & b & float & $[0.05, 0.999]$\\
\cline{2-5}
& XGBoostCost & - & - & -\\
\cline{2-5}
& LogRegCost & C & float & $[1e-6, 1e3]$\\
\cline{2-5}
& \multirow{3}{*}{DecisionTreeCost} & max\_depth & int & $[2, 10]$\\
& & min\_samples\_split & int & $[2, 10]$\\
& & min\_samples\_leaf & int & $[1, 5]$\\
\cline{2-5}
& \multirow{4}{*}{RandomForestCost} & n\_estimators & int & $[20, 200]$\\
& & max\_depth & int & $[2, 10]$\\
& & min\_samples\_split & int & $[2, 10]$\\
& & min\_samples\_leaf & int & $[1, 5]$\\

\midrule[1.5pt]

\end{tabular}
\end{adjustbox}
\end{table}

\begin{table}[!ht]
\footnotesize
\centering
\ContinuedFloat
\caption{Hyperparameters for different methods(continued)}
\begin{adjustbox}{width=\textwidth}
\begin{tabular}{lllcl}

\toprule[2pt]
\textbf{Category} & \textbf{Algorithm} & \textbf{Hyperparameter} & \textbf{Type} & \textbf{Range/Values} \\
\midrule[1.5pt]

\multirow{25}{3cm}{Ensemble-based Methods} 
& \multirow{2}{*}{SelfPacedEnsemble} & n\_estimators & int & $[20, 200]$ \\
& & k\_bins & int & $[2, 10]$ \\
\cline{2-5}
& \multirow{1}{*}{BalanceCascadeEnsemble} & n\_estimators & int & $[20, 200]$ \\
\cline{2-5}
& \multirow{4}{*}{BalancedRandomForest} & n\_estimators & int & $[20, 200]$ \\
& & max\_depth & int & $[2, 10]$ \\
& & min\_samples\_split & int & $[2, 10]$ \\
& & min\_samples\_leaf & int & $[1, 5]$\\
\cline{2-5}
& \multirow{1}{*}{EasyEnsemble} & n\_estimators & int & $[20, 200]$ \\
\cline{2-5}
& \multirow{2}{*}{RUSBoost} & n\_estimators & int & $[20, 200]$ \\
& & learning\_rate & float & $[0.5, 2]$\\
\cline{2-5}
& \multirow{1}{*}{UnderBagging} & n\_estimators & int & $[20, 200]$ \\
\cline{2-5}
& \multirow{2}{*}{OverBoost} & n\_estimators & int & $[20, 200]$ \\
& & learning\_rate & float & $[0.5, 2]$\\
\cline{2-5}
& \multirow{3}{*}{SMOTEBoost} & n\_estimators & int & $[20, 200]$ \\
& & k\_neighbors & int & $[3, 10]$\\
& & learning\_rate & float & $[0.5, 2]$\\
\cline{2-5}
& \multirow{1}{*}{OverBagging} & n\_estimators & int & $[20, 200]$ \\
\cline{2-5}
& \multirow{2}{*}{SMOTEBagging} & n\_estimators & int & $[20, 200]$ \\
& & k\_neighbors & int & $[3, 10]$\\
\cline{2-5}
& \multirow{2}{*}{AdaCost} & n\_estimators & int & $[20, 200]$ \\
& & learning\_rate & float & $[0.5, 1.5]$\\
\cline{2-5}
& \multirow{2}{*}{AdaUCost} & n\_estimators & int & $[20, 200]$ \\
& & learning\_rate & float & $[0.5, 1.5]$\\
\cline{2-5}
& \multirow{2}{*}{AsymBoost} & n\_estimators & int & $[20, 200]$ \\
& & learning\_rate & float & $[0.5, 1.5]$\\

\bottomrule[2pt]
\end{tabular}
\end{adjustbox}\\[5mm]
\end{table}

\begin{table}[!ht]
\renewcommand\arraystretch{1.1}
\centering
\caption{Open-source Python tools for the methods used in the paper.}
\label{T.python}
\begin{tabular}{l|c|c}
\toprule[2pt]
Category  & Approach &  Python Packages \\
\midrule[1pt]

\multirow{3}{*}{Data-level} 
& Over-sampling 
& imbalanced-learn\footnotemark[1], smote-variants\footnotemark[2] \\
& Under-sampling 
& imbalanced-learn \\
& Hybrid 
& imbalanced-learn \\

\midrule[1pt]

\multirow{2}{*}{Algorithm-level} 
& Cost-sensitive
& scikit-learn\footnotemark[3], XGBoost\footnotemark[5] \\
& Loss modification 
& gbdtCBL\footnotemark[4] \\

\midrule[1pt]

\multirow{2}{*}{Ensemble} 
& Resampling
& imbalanced-learn, imbalanced-ensemble\footnotemark[6]\\
& Cost-incorporated 
& imbalanced-ensemble \\
\bottomrule[2pt]
\end{tabular}
\end{table}
\footnotetext[1]{\url{https://imbalanced-learn.org/stable/index.html}}
\footnotetext[2]{\url{https://smote-variants.readthedocs.io/en/latest/index.html}}
\footnotetext[3]{\url{http://scikit-learn.org/stable/}}
\footnotetext[4]{\url{https://github.com/Luojiaqimath/ClassbalancedLoss4GBDT}}
\footnotetext[5]{\url{https://xgboost.readthedocs.io/en/stable/}}
\footnotetext[6]{\url{https://imbalanced-ensemble.readthedocs.io/en/latest/}}

\section{Supplementary Results}
\label{a.sup}

\begin{figure}[!ht]
    \centering
    \includegraphics[width=0.95\linewidth]{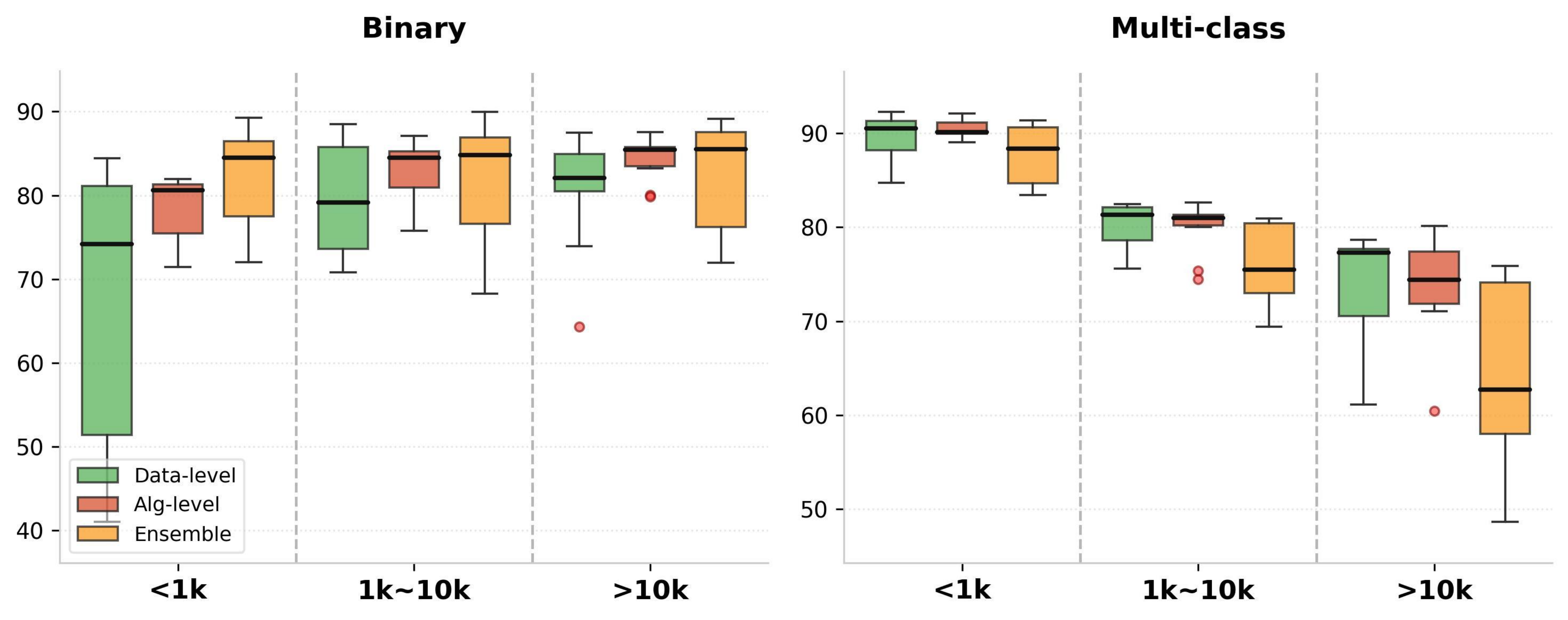}
    \caption{Family-level G-mean scores across sample size regimes. Each box shows the distribution of method performance within a family for each imbalance group. Results are reported separately for binary and multi-class tasks. Larger values indicate lower computational efficiency.}
    \label{size_g}
\end{figure}

\begin{figure}[!ht]
    \centering
    \includegraphics[width=0.95\linewidth]{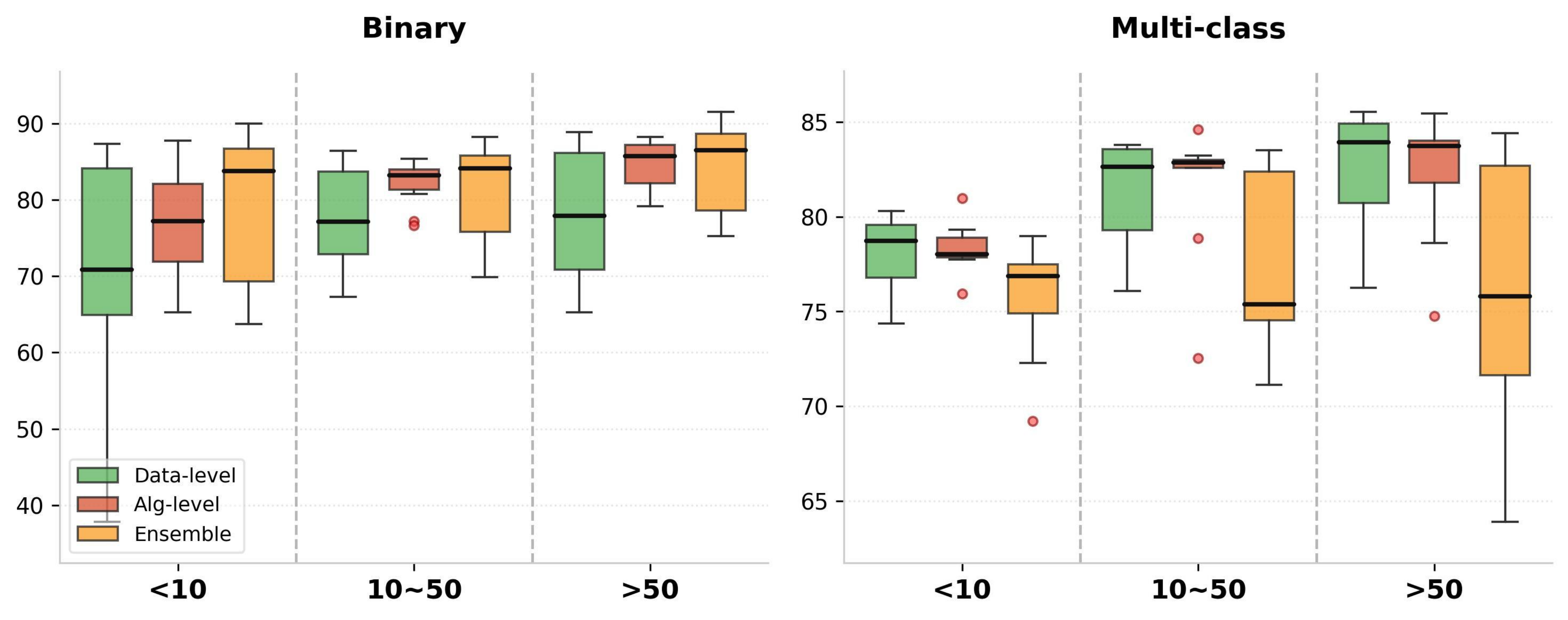}
    \caption{Family-level G-mean scores across feature dimensionality regimes. Each box shows the distribution of method performance within a family for each imbalance group. Results are reported separately for binary and multi-class tasks. Larger values indicate lower computational efficiency.}
    \label{dim_g}
\end{figure}

\begin{figure}[!ht]
    \centering
    \includegraphics[width=0.95\linewidth]{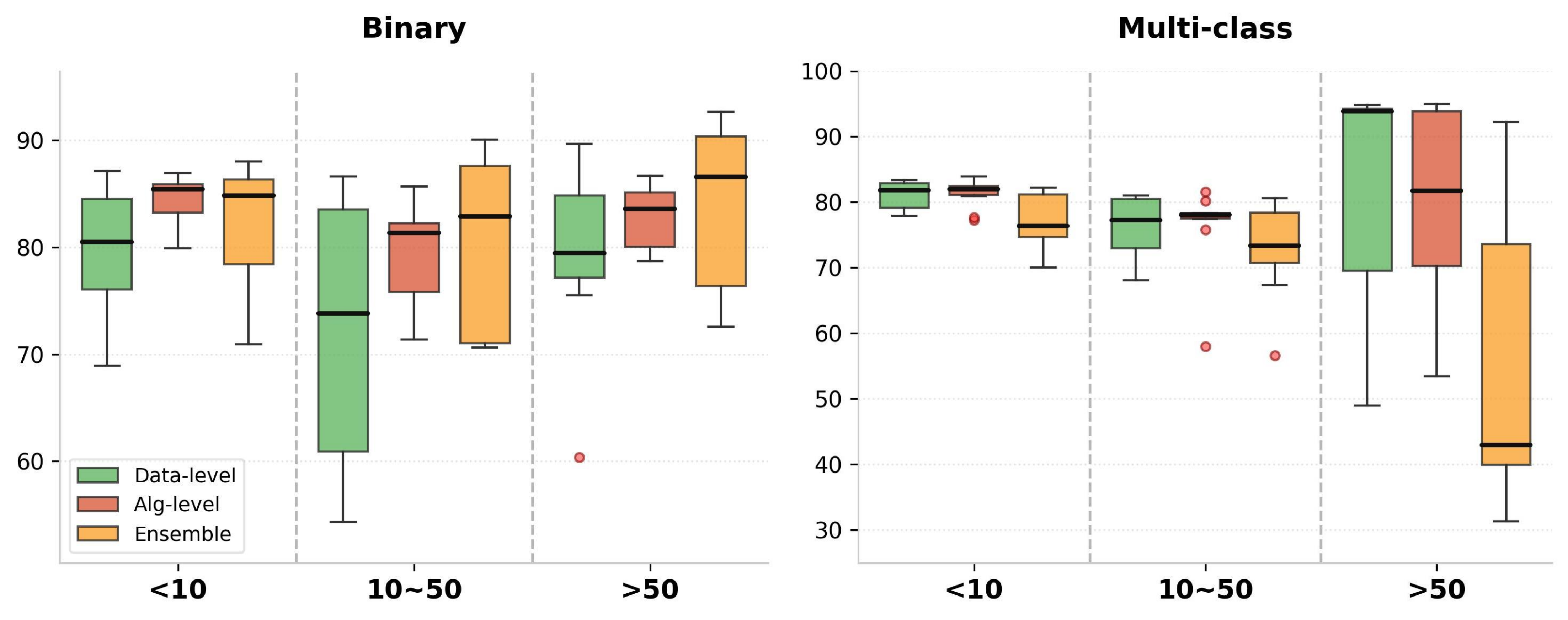}
    \caption{Family-level G-mean scores across imbalance severity regimes. Each box shows the distribution of method performance within a family for each imbalance group. Results are reported separately for binary and multi-class tasks. Larger values indicate lower computational efficiency.}
    \label{ir_g}
\end{figure}

We first report the complete results in terms of F1-score and G-mean in Tables~\ref{T.all_binary} and~\ref{T.all_multi}.
We also provide the G-mean results in Tables~\ref{SS_g}, \ref{FD_g}, and~\ref{IR_g}, with the corresponding visualizations shown in Figs.~\ref{size_g}, \ref{dim_g}, and~\ref{ir_g}. Table~\ref{T.abbr} summarizes the abbreviations of all evaluated methods. Detailed training times across different data regimes are reported in Tables~\ref{scal1time}, \ref{scal2time}, and~\ref{scal3time}.

\begin{table}[!ht]
\centering
\caption{Overall performance rankings of individual methods on binary classification tasks in terms of F1-score and G-mean.}
\label{T.all_binary}
\begin{adjustbox}{width=\linewidth}
\begin{tabular}{c|c|c||c|c}
\toprule[2pt]
\textbf{Rank} & \textbf{Method} & \textbf{F1-score} & \textbf{Method} & \textbf{G-mean score} \\
\midrule[1.5pt]
\multicolumn{5}{c}{Binary}\\
\midrule
1 & SelfPacedEnsemble & 73.88 $\pm$ 4.01 & UnderBagging & 89.32 $\pm$ 2.12\\
2 & XGBoostWCE & 72.96 $\pm$ 4.91 & BalancedRandomForest & 88.34 $\pm$ 2.38\\
3 & XGBoostASL & 72.62 $\pm$ 5.48 & SMOTEENN & 87.27 $\pm$ 3.21\\
4 & XGBoostAWE & 72.54 $\pm$ 4.57 & BalanceCascadeEnsemble & 87.20 $\pm$ 2.72\\
5 & BalanceCascadeEnsemble & 72.51 $\pm$ 4.55 & AdaUCost & 86.74 $\pm$ 3.43\\
6 & XGBoostACE & 72.07 $\pm$ 6.16 & EasyEnsemble & 86.70 $\pm$ 2.40\\
7 & SMOTEBagging & 70.93 $\pm$ 4.73 & XGBoostCost & 86.22 $\pm$ 3.39\\
8 & Lee & 70.53 $\pm$ 5.97 & SelfPacedEnsemble & 85.97 $\pm$ 3.14\\
9 & XGBoostFL & 70.14 $\pm$ 5.84 & SMOTEIPF & 85.00 $\pm$ 3.58\\
10 & XGBoostCost & 69.90 $\pm$ 4.90 & SMOTETomek & 84.92 $\pm$ 3.99\\
11 & BorderlineSMOTE & 69.71 $\pm$ 5.72 & SMOTE & 84.89 $\pm$ 3.98\\
12 & XGBoostCBE & 69.58 $\pm$ 5.14 & LogRegCost & 84.68 $\pm$ 3.16\\
13 & PolyfitSMOTE & 69.12 $\pm$ 6.28 & XGBoostAWE & 84.61 $\pm$ 3.84\\
14 & SMOTE & 67.95 $\pm$ 6.46 & OverBoost & 84.58 $\pm$ 4.62\\
15 & OverBagging & 67.86 $\pm$ 4.70 & BorderlineSMOTE & 84.48 $\pm$ 3.82\\
16 & RandomForestCost & 67.83 $\pm$ 4.83 & DecisionTreeCost & 84.40 $\pm$ 3.55\\
17 & SMOTETomek & 67.75 $\pm$ 6.05 & XGBoostASL & 83.87 $\pm$ 4.99\\
18 & SMOTEIPF & 65.64 $\pm$ 5.48 & XGBoostWCE & 83.66 $\pm$ 3.86\\
19 & SMOTEENN & 65.50 $\pm$ 5.07 & SMOTEBoost & 83.56 $\pm$ 4.28\\
20 & AdaCost & 65.19 $\pm$ 5.83 & RandomForestCost & 82.73 $\pm$ 4.61\\
21 & AsymBoost & 65.19 $\pm$ 5.83 & Lee & 82.37 $\pm$ 4.57\\
22 & SMOTEBoost & 64.48 $\pm$ 5.53 & XGBoostACE & 80.61 $\pm$ 5.88\\
23 & AdaUCost & 64.33 $\pm$ 4.45 & SMOTEBagging & 80.61 $\pm$ 3.86\\
24 & OverBoost & 63.79 $\pm$ 5.43 & CondensedNearestNeighbour & 79.87 $\pm$ 5.09\\
25 & NeighbourhoodCleaningRule & 63.34 $\pm$ 12.74 & PolyfitSMOTE & 77.80 $\pm$ 5.34\\
26 & SMOBD & 62.35 $\pm$ 10.62 & InstanceHardnessThreshold & 77.58 $\pm$ 14.30\\
27 & EditedNearestNeighbours & 61.73 $\pm$ 17.58 & XGBoostFL & 77.37 $\pm$ 5.10\\
28 & UnderBagging & 61.55 $\pm$ 3.53 & XGBoostCBE & 77.05 $\pm$ 4.63\\
29 & DecisionTreeCost & 61.35 $\pm$ 4.82 & OverBagging & 75.81 $\pm$ 3.97\\
30 & AllKNN & 61.23 $\pm$ 16.80 & NearMiss & 74.76 $\pm$ 4.76\\
31 & \cellcolor{gray!30}XGBoost & 60.92 $\pm$ 12.79 & AsymBoost & 73.86 $\pm$ 5.66\\
32 & TomekLinks & 60.21 $\pm$ 13.09 & AdaCost & 73.86 $\pm$ 5.66\\
33 & CondensedNearestNeighbour & 59.71 $\pm$ 7.00 & NeighbourhoodCleaningRule & 72.84 $\pm$ 13.64\\
34 & OneSidedSelection & 59.20 $\pm$ 13.33 & SMOBD & 71.61 $\pm$ 10.74\\
35 & BalancedRandomForest & 57.07 $\pm$ 3.54 & ClusterCentroids & 71.41 $\pm$ 3.98\\
36 & LogRegCost & 55.30 $\pm$ 3.35 & RUSBoost & 71.41 $\pm$ 7.59\\
37 & InstanceHardnessThreshold & 54.90 $\pm$ 11.30 & AllKNN & 71.14 $\pm$ 18.36\\
38 & EasyEnsemble & 53.77 $\pm$ 3.45 & EditedNearestNeighbours & 71.05 $\pm$ 19.26\\
39 & RUSBoost & 50.79 $\pm$ 8.50 & \cellcolor{gray!30}XGBoost & 67.17 $\pm$ 13.36\\
40 & NearMiss & 37.77 $\pm$ 5.51 & TomekLinks & 66.66 $\pm$ 14.31\\
41 & ClusterCentroids & 36.71 $\pm$ 3.91 & OneSidedSelection & 65.95 $\pm$ 14.48\\
\bottomrule[2pt]
\end{tabular}
\end{adjustbox}
\end{table}

\begin{table}[!ht]
\centering
\caption{Overall performance rankings of individual methods on multi-class classification tasks in terms of F1-score and G-mean.}
\label{T.all_multi}
\begin{adjustbox}{width=\linewidth}
\begin{tabular}{c|c|c||c|c}
\toprule[2pt]
\textbf{Rank} & \textbf{Method} & \textbf{F1-score} & \textbf{Method} & \textbf{G-mean score} \\
\midrule[1.5pt]
\multicolumn{5}{c}{Multi-class}\\
\midrule
1 & SMOTE & 75.84 $\pm$ 1.45 & XGBoostCost & 83.91 $\pm$ 1.16\\
2 & XGBoostCost & 75.83 $\pm$ 1.57 & SMOTE & 83.63 $\pm$ 1.00\\
3 & SMOTETomek & 75.49 $\pm$ 1.45 & SMOTETomek & 83.44 $\pm$ 0.94\\
4 & BorderlineSMOTE & 75.49 $\pm$ 1.71 & BorderlineSMOTE & 83.23 $\pm$ 1.26\\
5 & Lee & 75.47 $\pm$ 1.52 & Lee & 83.08 $\pm$ 1.04\\
6 & TomekLinks & 75.02 $\pm$ 1.67 & TomekLinks & 82.60 $\pm$ 1.29\\
7 & XGBoostFL & 74.69 $\pm$ 1.76 & SMOTEIPF & 82.49 $\pm$ 1.11\\
8 & XGBoostCBE & 74.69 $\pm$ 1.64 & XGBoostFL & 82.15 $\pm$ 1.29\\
9 & \cellcolor{gray!30}XGBoost & 74.63 $\pm$ 1.61 & XGBoostCBE & 82.15 $\pm$ 1.19\\
10 & XGBoostWCE & 74.61 $\pm$ 1.31 & XGBoostWCE & 82.12 $\pm$ 0.95\\
11 & OneSidedSelection & 74.42 $\pm$ 1.92 & \cellcolor{gray!30}XGBoost & 82.09 $\pm$ 1.19\\
12 & SMOTEIPF & 74.15 $\pm$ 1.83 & PolyfitSMOTE & 82.08 $\pm$ 0.94\\
13 & PolyfitSMOTE & 73.92 $\pm$ 1.36 & OneSidedSelection & 81.99 $\pm$ 1.41\\
14 & SMOBD & 73.74 $\pm$ 1.56 & UnderBagging & 81.92 $\pm$ 0.87\\
15 & SMOTEBagging & 73.11 $\pm$ 1.34 & SMOBD & 81.76 $\pm$ 1.19\\
16 & XGBoostACE & 72.65 $\pm$ 2.76 & SelfPacedEnsemble & 81.43 $\pm$ 1.05\\
17 & XGBoostASL & 72.55 $\pm$ 3.25 & SMOTEBagging & 81.26 $\pm$ 1.08\\
18 & OverBagging & 72.14 $\pm$ 1.32 & RandomForestCost & 81.20 $\pm$ 1.25\\
19 & NeighbourhoodCleaningRule & 71.75 $\pm$ 3.01 & NeighbourhoodCleaningRule & 80.88 $\pm$ 2.15\\
20 & XGBoostAWE & 71.66 $\pm$ 3.15 & EditedNearestNeighbours & 80.84 $\pm$ 1.86\\
21 & EditedNearestNeighbours & 71.39 $\pm$ 2.78 & XGBoostACE & 80.78 $\pm$ 2.22\\
22 & UnderBagging & 71.10 $\pm$ 1.13 & XGBoostASL & 80.66 $\pm$ 2.56\\
23 & RandomForestCost & 70.65 $\pm$ 1.64 & BalanceCascadeEnsemble & 80.57 $\pm$ 1.16\\
24 & SelfPacedEnsemble & 70.45 $\pm$ 1.45 & BalancedRandomForest & 80.50 $\pm$ 1.36\\
25 & BalanceCascadeEnsemble & 69.28 $\pm$ 1.47 & OverBagging & 80.30 $\pm$ 1.07\\
26 & BalancedRandomForest & 68.03 $\pm$ 1.77 & XGBoostAWE & 79.94 $\pm$ 2.51\\
27 & AllKNN & 67.39 $\pm$ 2.66 & SMOTEENN & 78.46 $\pm$ 1.75\\
28 & CondensedNearestNeighbour & 67.00 $\pm$ 3.39 & AllKNN & 78.17 $\pm$ 2.00\\
29 & DecisionTreeCost & 64.90 $\pm$ 2.04 & CondensedNearestNeighbour & 78.10 $\pm$ 2.45\\
30 & SMOTEENN & 64.59 $\pm$ 2.42 & DecisionTreeCost & 77.60 $\pm$ 1.45\\
31 & NearMiss & 62.91 $\pm$ 1.78 & NearMiss & 76.79 $\pm$ 1.41\\
32 & InstanceHardnessThreshold & 61.63 $\pm$ 2.96 & InstanceHardnessThreshold & 76.38 $\pm$ 2.14\\
33 & LogRegCost & 60.11 $\pm$ 1.55 & ClusterCentroids & 75.78 $\pm$ 1.15\\
34 & AdaUCost & 59.79 $\pm$ 3.17 & AdaUCost & 74.19 $\pm$ 2.18\\
35 & AsymBoost & 59.55 $\pm$ 2.14 & LogRegCost & 74.12 $\pm$ 1.14\\
36 & SMOTEBoost & 59.54 $\pm$ 3.17 & OverBoost & 73.73 $\pm$ 1.98\\
37 & AdaCost & 59.48 $\pm$ 2.18 & SMOTEBoost & 73.49 $\pm$ 2.11\\
38 & OverBoost & 58.97 $\pm$ 2.65 & AsymBoost & 71.99 $\pm$ 1.68\\
39 & ClusterCentroids & 58.47 $\pm$ 1.62 & AdaCost & 71.96 $\pm$ 1.70\\
40 & EasyEnsemble & 55.06 $\pm$ 1.78 & EasyEnsemble & 70.88 $\pm$ 1.43\\
41 & RUSBoost & 50.44 $\pm$ 6.00 & RUSBoost & 66.77 $\pm$ 3.28\\
\bottomrule[2pt]
\end{tabular}
\end{adjustbox}
\end{table}

\begin{table}[!ht]
\centering
\caption{Top five methods in each sample size regime ranked by G-mean score for binary and multi-class tasks.}
\label{SS_g}
\begin{adjustbox}{width=\linewidth}
\begin{tabular}{c|c|c|c|c|c}
\toprule[2pt]
\textbf{Sample Size} & \textbf{Rank} & \textbf{Method} & \textbf{G-mean score} & \textbf{Method} & \textbf{G-mean score}\\
\midrule[1.5pt]
&&\multicolumn{2}{c|}{Binary}&\multicolumn{2}{c}{Multi-class}\\
\midrule
\multirow{5}{*}{$<$1k} & 1 & EasyEnsemble & 89.25 $\pm$ 4.94 & PolyfitSMOTE & 92.25 $\pm$ 1.79\\
& 2 & UnderBagging & 88.42 $\pm$ 4.95 &  XGBoostCost & 92.08 $\pm$ 1.95\\
& 3 & BalancedRandomForest & 86.96 $\pm$ 5.27 & SMOTE  & 91.81 $\pm$ 1.77\\
& 4 & BalanceCascadeEnsemble & 86.50 $\pm$ 6.09 & SMOTEIPF &91.70 $\pm$ 1.77\\
& 5 & SMOTEBoost & 84.80 $\pm$ 7.91 & UnderBagging & 91.35 $\pm$ 1.66\\
\midrule
\multirow{5}{*}{1k--10k} & 1 & UnderBagging & 89.94 $\pm$ 1.91 & XGBoostCost & 82.63 $\pm$ 1.16 \\
& 2 & BalancedRandomForest & 89.31 $\pm$ 1.71 & SMOTE & 82.44 $\pm$ 1.01\\
& 3 & SMOTEENN & 88.51 $\pm$ 2.67 & SMOTEIPF & 82.43 $\pm$ 0.87\\
& 4 & BalanceCascadeEnsemble & 88.26 $\pm$ 2.08 &BorderlineSMOTE & 82.26 $\pm$ 0.96\\
& 5 & XGBoostCost & 87.11 $\pm$ 3.02 & SMOTETomek & 82.26 $\pm$ 0.96 \\
\midrule
\multirow{5}{*}{$>$10k} & 1 & UnderBagging & 89.12 $\pm$ 0.73 & XGBoostCost & 80.14 $\pm$ 0.48\\
& 2 & AdaUCost & 88.44 $\pm$ 1.07 & SMOTETomek & 78.65 $\pm$ 0.34\\
& 3 & BalancedRandomForest & 87.72 $\pm$ 1.12 & SMOTE & 78.59 $\pm$ 0.34\\
& 4 & OverBoost & 87.57 $\pm$ 1.82 & BorderlineSMOTE & 78.41 $\pm$ 0.43\\
& 5 & XGBoostCost & 87.56 $\pm$ 1.48 & SMOBD & 78.05 $\pm$ 0.37\\

\bottomrule[2pt]
\end{tabular}
\end{adjustbox}
\end{table}

\begin{table}[!ht]
\centering
\caption{Top five methods in each feature dimensionality regime ranked by G-mean score for binary and multi-class tasks.}
\label{FD_g}
\begin{adjustbox}{width=\linewidth}
\begin{tabular}{c|c|c|c|c|c}
\toprule[2pt]
\textbf{Feature Dimension} & \textbf{Rank} & \textbf{Method} & \textbf{G-mean score}& \textbf{Method} & \textbf{G-mean score}\\
\midrule[1.5pt]
&&\multicolumn{2}{c|}{Binary}&\multicolumn{2}{c}{Multi-class}\\
\midrule
\multirow{5}{*}{$<$10} & 1 & BalancedRandomForest & 89.97 $\pm$ 2.39 & XGBoostCost & 82.98 $\pm$ 1.86\\
& 2 & UnderBagging & 89.28 $\pm$ 4.09 & SMOTE & 82.29 $\pm$ 1.59\\
& 3 & EasyEnsemble & 87.83 $\pm$ 3.58 & SMOTEIPF & 82.09 $\pm$ 1.47\\
& 4 & LogRegCost & 87.77 $\pm$ 3.20 & SMOTETomek & 81.86 $\pm$ 1.45\\
& 5 & SMOTEENN & 87.34 $\pm$ 4.72 &  Tomek & 81.61 $\pm$ 2.29\\
\midrule
\multirow{5}{*}{10--50} & 1 & UnderBagging & 88.26 $\pm$ 1.80 & XGBoostCost & 84.60 $\pm$ 0.90\\
& 2 & BalancedRandomForest & 87.80 $\pm$ 1.68 & BorderlineSMOTE & 83.77 $\pm$ 0.68\\
& 3 & SMOTEENN & 86.40 $\pm$ 2.38 & PolyfitSMOTE & 83.75 $\pm$ 0.74\\
& 4 & BalanceCascadeEnsemble & 86.15 $\pm$ 2.23 & SMOTEIPF & 83.69 $\pm$ 0.72\\
& 5 & AdaUCost & 85.82 $\pm$ 2.46 & SMOTE & 83.62 $\pm$ 0.84\\
\midrule
\multirow{5}{*}{$>$50} & 1 & UnderBagging & 91.51 $\pm$ 2.46 & SMOTE & 85.51 $\pm$ 0.41\\
& 2 & BalanceCascadeEnsemble & 91.07 $\pm$ 3.32 & XGBoostCost & 85.45 $\pm$ 0.58\\
& 3 & SelfPacedEnsemble & 89.20 $\pm$ 3.82 & SMOTETomek & 85.45 $\pm$ 0.44\\
& 4 & SMOTEENN & 88.89 $\pm$ 4.10 & BorderlineSMOTE & 85.25 $\pm$ 0.55\\
& 5 & EasyEnsemble & 88.65 $\pm$ 3.23 & PolyfitSMOTE & 84.95 $\pm$ 0.55\\

\bottomrule[2pt]
\end{tabular}
\end{adjustbox}
\end{table}

\begin{table}[!ht]
\centering
\caption{Top five methods in each imbalance severity regime ranked by G-mean score for binary and multi-class tasks.}
\label{IR_g}
\begin{adjustbox}{width=\linewidth}
\begin{tabular}{c|c|c|c|c|c}
\toprule[2pt]
\textbf{Imbalance Ratio} & \textbf{Rank} & \textbf{Method} & \textbf{G-mean score} & \textbf{Method} & \textbf{G-mean score}\\
\midrule[1.5pt]
&&\multicolumn{2}{c|}{Binary}&\multicolumn{2}{c}{Multi-class}\\
\midrule
\multirow{5}{*}{$<$10} & 1 & UnderBagging & 88.00 $\pm$ 1.79 & XGBoostCost & 83.92 $\pm$ 1.20\\
& 2 & BalancedRandomForest & 87.34 $\pm$ 1.74 & SMOTE & 83.32 $\pm$ 1.09\\
& 3 & SMOTEENN & 87.13 $\pm$ 1.82 & SMOTETomek & 83.12 $\pm$ 1.03\\
& 4 & XGBoostCost & 86.90 $\pm$ 2.16 & PolyfitSMOTE & 83.06 $\pm$ 1.00\\
& 5 & AdaUCost & 86.72 $\pm$ 2.07 & Lee & 82.98 $\pm$ 1.09\\
\midrule
\multirow{5}{*}{10--50} & 1 & UnderBagging & 90.05 $\pm$ 3.23 & XGBoostCost & 81.61 $\pm$ 1.04\\
& 2 & BalancedRandomForest & 88.98 $\pm$ 3.44 & SMOTEIPF & 81.01 $\pm$ 0.67\\
& 3 & BalanceCascadeEnsemble & 88.80 $\pm$ 4.92 & SMOTE & 80.87 $\pm$ 0.49\\
& 4 & EasyEnsemble & 87.62 $\pm$ 3.70 & SMOTETomek & 80.62 $\pm$ 0.36\\
& 5 & SelfPacedEnsemble & 87.05 $\pm$ 4.67 & SelfPacedEnsemble & 80.60 $\pm$ 0.71\\
\midrule
\multirow{5}{*}{$>$50} & 1 & UnderBagging & 92.62 $\pm$ 1.67 & XGBoostCost & 95.01 $\pm$ 0.56\\
& 2 & EasyEnsemble & 91.27 $\pm$ 1.66 & EditedNearestNeighbours & 94.80 $\pm$ 0.42\\
& 3 & BalanceCascadeEnsemble & 90.48 $\pm$ 2.51 & XGBoost & 94.56 $\pm$ 0.32\\
& 4 & BalancedRandomForest & 90.33 $\pm$ 1.90 & SMOTETomek & 94.50 $\pm$ 0.53\\
& 5 & SMOTEENN & 89.65 $\pm$ 3.31 & OneSidedSelection & 94.41 $\pm$ 0.73\\

\bottomrule[2pt]
\end{tabular}
\end{adjustbox}
\end{table}

\begin{table}[!ht]
\centering
\caption{Performance of methods that support missing values. We report G-mean scores for both binary and multi-class tasks. Methods without native support are excluded.}
\label{T.gmiss}
\begin{tabular}{c|c|c}
\toprule[2pt]
\textbf{Method} & \textbf{Binary} & \textbf{Multi-class}\\
\midrule[1.5pt]
\multicolumn{3}{c}{Algorithm-level}\\
\midrule
XGBoostASL & 83.87 $\pm$ 4.99 & 80.66 $\pm$ 2.56\\
XGBoostACE & 80.61 $\pm$ 5.88 & 80.78 $\pm$ 2.22\\
XGBoostAWE& 84.61 $\pm$ 3.84 & 79.94 $\pm$ 2.51\\
XGBoostFL & 77.37 $\pm$ 5.10 & 82.15 $\pm$ 1.29\\
XGBoostWCE & 83.66 $\pm$ 3.86 & 82.12 $\pm$ 0.95\\
XGBoostCBE & 77.05 $\pm$ 4.63 & 82.15 $\pm$ 1.19\\
XGBoostCost & 86.22 $\pm$ 3.39 & 83.91 $\pm$ 1.16\\
DecisionTreeCost & 84.40 $\pm$ 3.55 & 77.60 $\pm$ 1.45\\
RandomForestCost & 82.73 $\pm$ 4.61 & 81.20 $\pm$ 1.25\\
\midrule[1.5pt]
\multicolumn{3}{c}{Ensemble}\\
\midrule
SelfPacedEnsemble & 85.97 $\pm$ 3.14 & 81.43 $\pm$ 1.05\\
BalanceCascadeEnsemble & 87.20 $\pm$ 2.72 & 80.57 $\pm$ 1.16\\
UnderBagging & 89.32 $\pm$ 2.12 & 81.92 $\pm$ 0.87\\
OverBagging & 75.81 $\pm$ 3.97 & 80.30 $\pm$ 1.07\\
\bottomrule[2pt]
\end{tabular}
\end{table}


\begin{table}[ht]
\centering
\caption{Method abbreviations used in figures.}
\label{T.abbr}
\begin{adjustbox}{width=\textwidth}
\begin{tabular}{l|c|l|l|c|l}
\toprule
\textbf{Family} & \textbf{Abbreviation} & \textbf{Method} & \textbf{Family} & \textbf{Abbreviation} & \textbf{Method} \\
\midrule
Baseline & XGB & XGBoost & \multirow{7}{*}{Alg.-level} & FL & XGBoostFL \\
\cline{1-3}
\multirow{17}{*}{Data-level} & TL & TomekLinks & & WCE & XGBoostWCE \\
& ENN & EditedNearestNeighbours & & CBE & XGBoostCBE \\
& NCR & NeighbourhoodCleaningRule & & XC & XGBoostCost \\
& IHT & InstanceHardnessThreshold & & LRC & LogRegCost \\
& CC & ClusterCentroids & & DTC & DecisionTreeCost \\
& CNN & CondensedNearestNeighbour & & RFC & RandomForestCost \\
\cline{4-6}
& AK & AllKNN & \multirow{13}{*}{Ensemble} & SPE & SelfPacedEnsemble \\
& NM & NearMiss & & BCE & BalanceCascadeEnsemble \\
& OSS & OneSidedSelection & & BRF & BalancedRandomForest \\
& SMO & SMOTE & & EE & EasyEnsemble \\
& BLS & BorderlineSMOTE & & RUS & RUSBoost \\
& PFS & PolyfitSMOTE & & UBA & UnderBagging \\
& SIPF & SMOTEIPF & & OBO & OverBoost \\
& Lee & Lee & & SBO & SMOTEBoost \\
& SBD & SMOBD & & OBA & OverBagging \\
& SENN & SMOTEENN & & SBA & SMOTEBagging \\
& ST & SMOTETomek & & AC & AdaCost \\
\cline{1-3}
\multirow{3}{*}{Alg.-level} & ASL & XGBoostASL & & AUB & AdaUCost/AdaUBoost \\
& ACE & XGBoostACE & & ASB & AsymBoost \\
& AWE & XGBoostAWE & & & \\
\bottomrule
\end{tabular}
\end{adjustbox}
\end{table}

\begin{table}[!ht]
\centering
\caption{Running time comparison across different sample sizes. Runtime values are reported as $\log_{10}(\text{seconds})$ and averaged over five independent runs. We report family-level average, minimum, and maximum runtime. Larger values indicate lower computational efficiency.}
\label{scal1time}
\begin{adjustbox}{width=\linewidth}
\begin{tabular}{c|c|c|c|c|c}
\toprule[2pt]
\multicolumn{2}{c|}{} & \textbf{\large XGBoost} & \textbf{\large Data-level} & \textbf{\large Alg.-level} & \textbf{\large Ensemble}\\
\midrule[1.5pt]
\multirow{4}{*}{1k Samples} & Rank & --& 2& 1&3 \\
\cline{2-6}
& Avg. &-1.24 &-0.71 & -1.20&-0.57 \\
\cline{2-6}
& Min &-1.24 &-1.36 &-2.80 &-1.14 \\
\cline{2-6}
& Max &-1.24 &0.12  &-0.62 & -0.23\\
\midrule[1pt]
\multirow{4}{*}{100k Samples} & Rank & --&3 &1 &2 \\
\cline{2-6}
& Avg. &-0.40 &2.90& -0.27& 1.02\\
\cline{2-6}
& Min &-0.40 &-0.17 &-1.42 &0.04 \\
\cline{2-6}
& Max &-0.40 &4.13 & -0.01 & 1.33 \\
\bottomrule[2pt]
\end{tabular}
\end{adjustbox}
\end{table}

\begin{table}[!ht]
\centering
\caption{Running time comparison across different feature dimensions. Runtime values are reported as $\log_{10}(\text{seconds})$ and averaged over five independent runs. We report family-level average, minimum, and maximum runtime. Larger values indicate lower computational efficiency.}
\label{scal2time}
\begin{adjustbox}{width=\linewidth}
\begin{tabular}{c|c|c|c|c|c}
\toprule[2pt]
\multicolumn{2}{c|}{} & \textbf{\large XGBoost} & \textbf{\large Data-level} & \textbf{\large Alg.-level} & \textbf{\large Ensemble}\\
\midrule[1.5pt]
\multirow{4}{*}{50 Features} & Rank & --& 2& 1&3 \\
\cline{2-6}
& Avg. &-0.58 &0.51 & -0.48&0.70 \\
\cline{2-6}
& Min &-0.58 &-0.58 &-1.64 &-0.29 \\
\cline{2-6}
& Max &-0.58 &1.39  &-0.33 & 1.10\\
\midrule[1pt]
\multirow{4}{*}{500 Features} & Rank & --&2 &1 &3 \\
\cline{2-6}
& Avg. &0.26 &1.50 & 0.33& 1.66\\
\cline{2-6}
& Min &0.26 &0.03 &-0.53 &-0.17 \\
\cline{2-6}
& Max &0.26 &2.55 & 0.64 & 2.17 \\
\bottomrule[2pt]
\end{tabular}
\end{adjustbox}
\end{table}

\begin{table}[!ht]
\centering
\caption{Running time comparison across different class numbers. Runtime values are reported as $\log_{10}(\text{seconds})$ and averaged over five independent runs. We report family-level average, minimum, and maximum runtime. Larger values indicate lower computational efficiency.}
\label{scal3time}
\begin{adjustbox}{width=\linewidth}
\begin{tabular}{c|c|c|c|c|c}
\toprule[2pt]
\multicolumn{2}{c|}{} & \textbf{\large XGBoost} & \textbf{\large Data-level} & \textbf{\large Alg.-level} & \textbf{\large Ensemble}\\
\midrule[1.5pt]
\multirow{4}{*}{2 Classes} & Rank & --& 3& 1&2 \\
\cline{2-6}
& Avg. &-1.35 &0.24 & -1.15 &-0.09 \\
\cline{2-6}
& Min &-1.35 &-1.29 &-2.12 &-0.88 \\
\cline{2-6}
& Max &-1.35 &1.26  &-0.66 & 0.33\\
\midrule[1pt]

\multirow{4}{*}{20 Classes} & Rank & --& 3&1 &2 \\
\cline{2-6}
& Avg. &0.34 &1.55 & 0.11& 0.81\\
\cline{2-6}
& Min &0.34 &0.16 &-0.67 &0.05 \\
\cline{2-6}
& Max &0.34 &2.63 & 0.50 &1.49 \\
\bottomrule[2pt]
\end{tabular}
\end{adjustbox}
\end{table}

\bibliographystyle{elsarticle-num} 
\bibliography{references}

\end{document}